\pdfoutput=1

\documentclass[11pt]{article}

\usepackage[final]{acl}

\usepackage{times}
\usepackage{latexsym}
\usepackage{enumitem}

\usepackage[T1]{fontenc}

\usepackage[utf8]{inputenc}

\usepackage{microtype}

\usepackage{inconsolata}

\usepackage{graphicx}
\usepackage{booktabs}
\usepackage[export]{adjustbox}

\usepackage{amsmath,amsfonts,bm}

\def\eqref#1{equation~\ref{#1}}

\def\1{\bm{1}}

\def\rvtheta{{\bm{\theta}}}

\def\vx{{\bm{x}}}
\def\vy{{\bm{y}}}

\DeclareMathAlphabet{\mathsfit}{\encodingdefault}{\sfdefault}{m}{sl}
\SetMathAlphabet{\mathsfit}{bold}{\encodingdefault}{\sfdefault}{bx}{n}

\def\sD{{\mathbb{D}}}

\newcommand{\sigmoid}{\sigma}

\def\ftparam{\textcolor{blue}{{\pi}_{\rvtheta^0}}}
\def\genparam{\textcolor{red}{{\rho}}}

\def\policyparam{\textcolor{blue}{{\pi}_\rvtheta}}
\def\policyparamt{\textcolor{blue}{{\pi}_{\rvtheta^t}}}

\usepackage{multirow}
\usepackage{url}
\usepackage{xcolor} 
\usepackage{ulem}

\title{Towards Adapting Open-Source Large Language Models for Expert-Level Clinical Note Generation}

\author{
Hanyin Wang$^{1,2}$,
Chufan Gao$^2$, Bolun Liu$^1$, Qiping Xu$^1$, Guleid Hussein$^1$, \\ \textbf{Mohamad El Labban$^1$, Kingsley Iheasirim$^1$,  Hariprasad Korsapati$^1$,} \\ \textbf{Chuck Outcalt $^3$, Jimeng Sun$^{2,4}$}\\
\\
$^1$ Mayo Clinic Health System, $^2$ School of Computing and Data Science, UIUC \\ $^3$ Mayo Clinic Rochester, $^4$ Carle Illinois College of Medicine, UIUC\\
\texttt{wang.hanyin@mayo.edu, jimeng@illinois.edu}
}

\begin{document}

\maketitle
\begin{abstract}
Proprietary Large Language Models (LLMs) such as GPT-4 and Gemini have demonstrated promising capabilities in clinical text summarization tasks. However, due to patient data privacy concerns and computational costs, many healthcare providers prefer using small, locally-hosted models over external generic LLMs.
This study presents a comprehensive domain- and task-specific adaptation process for the open-source LLaMA-2 13 billion parameter model, enabling it to generate high-quality clinical notes from outpatient patient-doctor dialogues. 
Our process incorporates continued pretraining, supervised fine-tuning, and reinforcement learning from both AI and human feedback. We introduced a new approach, {\it DistillDirect}, for performing on-policy reinforcement learning with Gemini 1.0 Pro as the teacher model. Our resulting model, LLaMA-Clinic, can generate clinical notes comparable in quality to those authored by physicians. In a blinded physician reader study, the majority (92.8\%) of individual evaluations rated the notes generated by LLaMA-Clinic as ``acceptable'' or higher across three criteria: real-world readiness, completeness, and accuracy. In the more challenging ``Assessment and Plan'' section,  LLaMA-Clinic matched physician-authored notes in real-world readiness score.
We highlight key considerations for future clinical note-generation tasks, emphasizing the importance of pre-defining a ``\textbf{best practice}'' note format, rather than relying on LLMs to determine this for clinical practice. \footnote{Our code and data are available at \url{https://github.com/hanyin88/llama-clinic}.}
\end{abstract}

\section{Introduction}\label{introduction}
Recent advancements in LLMs have transformed the field of natural language processing (NLP). However, the application of LLMs in the medical domain is still in its early stages ~\cite{he2023survey,zhou2023survey}. Proprietary LLMs, such as GPT-4 and Med-PaLM, have demonstrated impressive capabilities in medical knowledge and clinical NLP tasks \cite{nori2023capabilities, singhal2023expertlevel, van2023clinical}. However, most proprietary LLMs have limited flexibility for domain-specific fine-tuning, primarily due to restricted access to model weights. Additionally, proprietary LLMs raise several concerns pertinent to the healthcare sector, including HIPAA compliance, data security, cost, and transparency of training data \cite{marks2023ai, clusmann2023future, Adimi2023, nyt_article_2023}.

The emergence of powerful open-source LLMs has opened up opportunities for domain-specific fine-tuning within the clinical field, yielding promising results \cite{han2023medalpaca, wang2023huatuo, li2023chatdoctor, wu2023pmcllama,toma2023clinical, chen2023meditron}. However, most research on open-source models has concentrated on medical knowledge injection rather than practical applications in real-world clinical workflow such as clinical note generation. 

In this work, we address a practical question clinicians face in their everyday routine: \textit{How can we best adapt an open-source LLM for the specific use case of clinical note generation?} Clinical note documentation represents a significant burden for healthcare practitioners \cite{ammenwerth2009time} and appears to be a natural application for LLMs, given their remarkable generative capabilities \cite{lee2023benefits}. Recent research on LLMs for clinical text summarization found that LLM-generated outputs are preferred over human summaries for their completeness and accuracy \cite{van2023clinical}.

\begin{figure*}[!ht]
    \captionsetup{labelfont=bf={bf},skip=12pt}
    \centering
    \includegraphics[width=1\linewidth]{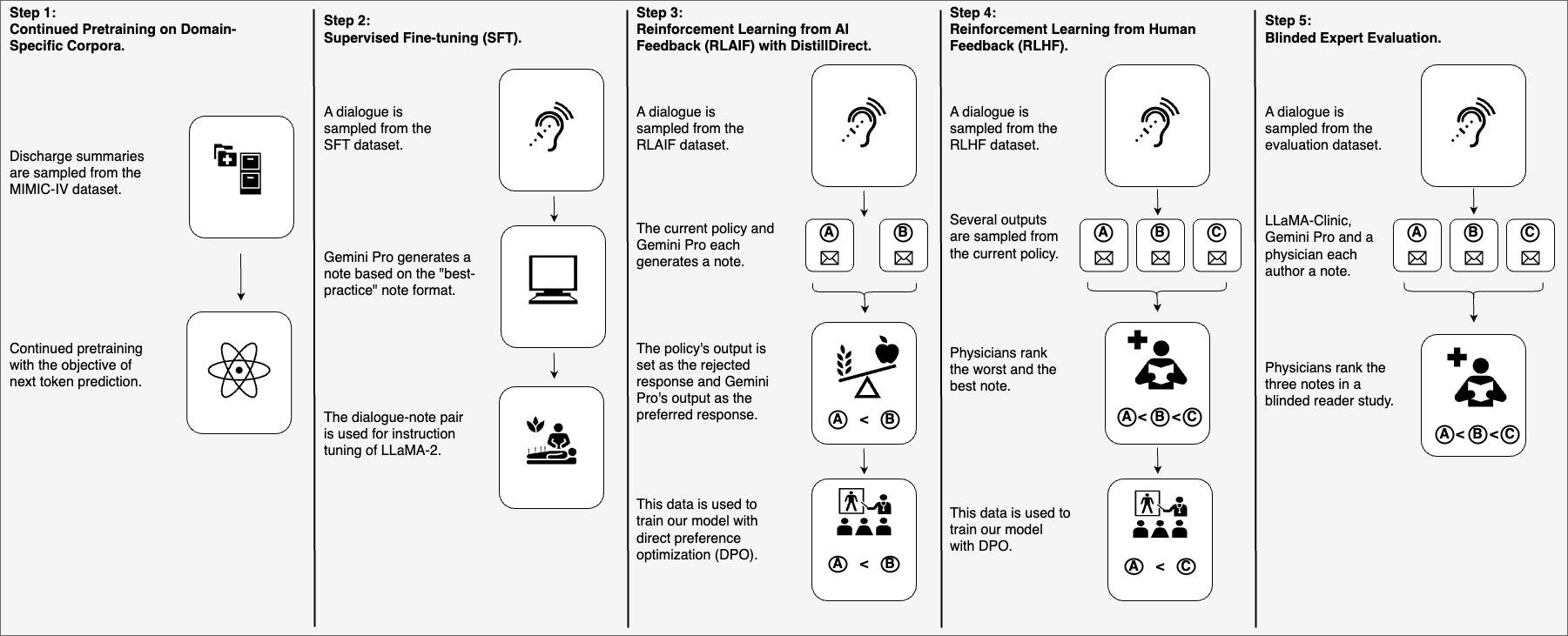}
    \caption{\textbf{Overview of Study Design.} We conducted a comprehensive domain- and task-specific adaptation process for the LLaMA-2-13B model. This process included continued pretraining, supervised fine-tuning, and reinforcement learning from AI and human feedback. Finally, we evaluated our model's outputs against those created by physicians and Gemini Pro through a blinded expert evaluation. We used Gemini 1.0 Pro as the teacher model in this study. }
    \label{fig:overview}
\end{figure*}

The 2023 ACL ClinicalNLP and CLEFImage workshops explored the generation of clinical notes from patient-doctor conversations using the newly released Ambient Clinical Intelligence Benchmark (ACI-BENCH) dataset \cite{abacha2023overview, yim2023overview}. The most notable results were achieved using GPT-4 along with few-shot in-context learning \cite{giorgi2023wanglab, van2023clinical}. However, these initial exploratory studies for clinical note generation with LLMs revealed significant limitations: 
\begin{itemize}[leftmargin=*]
    \item {\bf Issue with metrics:} The evaluation relied solely on automatic metrics of lexical similarity to the ``reference'' notes, such as ROUGE scores, which may not accurately reflect human preferences \cite{moramarco2022human, van2023clinical}.
    \item {\bf Variation in ground-truth quality:} There was considerable variation in the quality, format, and style of ``reference notes'', suggesting that a note similar to a reference note might not necessarily be of high quality for real-world clinical applications.
    \item {\bf Limited fine-tuning datasets:} Previous studies conducted only limited supervised fine-tuning (SFT) of open-source LLMs using small training datasets, thereby not fully exploring their potential for domain- and task-specific adaptation \cite{abacha2023overview, yim2023overview, van2023clinical}.
    \item {\bf Lack of advanced training strategy:} The potential of data augmentation and reinforcement learning remains unexplored.
\end{itemize}
   
In this study, we revisit the task of outpatient note generation, focusing on adapting an open-source LLM---the LLaMA-2 13 billion parameter model. We thoroughly evaluated techniques for domain- and task-specific adaptation, ranging from continued pretraining and SFT to reinforcement learning informed from both AI and human feedback (Figure \ref{fig:overview}). This work makes several specific contributions:
\begin{enumerate}[leftmargin=*]
    \item {\bf  Clinical LLM fine-tuning playbook:} Our relatively compact, open-source LLM demonstrated expert-level performance, achieving the same quality as physician-authored notes in the ``Assessment and Plan" section. We provide a comprehensive guide for healthcare organizations to fine-tune local LLMs on their own data. 
    \item {\bf DistillDirect:} We are among the first groups to explore the role of reinforcement learning in clinical note generation. We proposed  {\it DistillDirect}, a strategy to ensure on-policy learning during direct preference optimization (DPO) for model distillation.
    \item {\bf Open datasets:}  We have made our newly created synthetic clinic dialogue-note dataset and physician feedback dataset publicly available. 
    \item {\bf Key considerations for future clinical note-generation tasks:} We underscored the importance of pre-defining a ``best practice'' note format, rather than relying on LLMs to determine this for clinical practice. We also recommend a workflow for physicians to work with AI-generated notes in real-world practice, ensuring physicians' full oversight and ultimate accountability for the AI-generated content.
\end{enumerate}

\section{Background and Related Work}\label{background}
\subsection{Problem Formation}
Given a recorded dialogue from a patient-doctor clinic encounter, we task LLMs to generate a high-quality outpatient note akin to one written by a clinician. This scenario is becoming increasingly prevalent due to the rising popularity of ambient AI scribe products \cite{yim2023aci, barr2024preparing}. Our research focus is on generating the ``Subjective'' and ``Assessment and Plan'' sections of outpatient notes. This decision is based on feedback from our physician coauthors, which indicates that discussing all details of the ``Objective'' section, such as physical examination results, is impractical during real-world clinical encounters.
Furthermore, much of the objective data, including lab and imaging results, are directly integrated into Electronic Medical Records (EMR), making clinical notes generation for those sections easy and sometimes unnecessary. 
Our research is also related to prior work on generating clinical assessments within notes~\cite{yang2020generating} and on generating doctor-patient conversations~\cite{wang2024notechat}.

\subsection{ACI-BENCH} 
ACI-BENCH represents the largest clinic dialogue-note dataset publicly available to date, comprising 207 cases \cite{yim2023aci}. The dataset's dialogues were crafted by a team with medical expertise, and its clinical notes were initially generated using an automatic note generation system, then reviewed and revised by domain experts \cite{yim2023aci}. The ACI-BENCH dataset was previously utilized to benchmark the performance of outpatient note generation systems using automatic metrics that evaluate lexical similarity \cite{abacha2023overview, yim2023overview}. Importantly, we observed notable variation in the format, style, and quality of the ``reference notes'' within ACI-BENCH, especially in the section of ``Assessment and Plan''. While this diversity mirrors the reality of clinical practice, where different doctors may produce vastly different notes, it presents a challenge to use these notes as a ``gold standard'' for training an LLM to replicate.

\subsection{Distilled DPO}
DPO begins by collecting a preference dataset $\sD = \{(\vx_i, \vy_i^+, \vy_i^-)\}_{i=1}^{N} $, where for each prompt $\vx$, there is a preferred answer $\vy^+$ and a rejected answer $\vy^-$ \cite{rafailov2023direct}. Following the notations as in \cite{guo2024direct}, DPO optimizes the language model (target policy $\policyparam$) using the following loss function: 
\begin{equation}
        \hspace{-0.4cm}
         - \log\sigmoid\left( \beta\log\frac{\policyparam(\vy^+|\vx)\ftparam(\vy^-|\vx)}{\ftparam(\vy^+|\vx)\policyparam(\vy^-|\vx)} \right)
    \label{eq:background:dpo_objective}
    \end{equation}

Here, $\ftparam$ represents the SFT baseline used as a reference, $\sigmoid$ denotes the logistic function, and $\beta$ is introduced as a scalar hyperparameter. 

Given the significant time and financial costs associated with collecting preference data, utilizing pre-collected preference datasets, such as those employed in distilled DPO, is a common practice. Distilled DPO involves generating a collection of responses for each prompt from various LLMs (Figure \ref{fig:distilldirect}A) \cite{tunstall2023zephyr}. These responses are then evaluated by a teacher model (e.g., GPT-4) to provide preference feedback. Applying reinforcement learning with AI feedback (RLAIF) with distilled DPO has yielded encouraging outcomes \cite{tunstall2023zephyr}. However, distilled DPO's reliance on a pre-collected preference dataset renders it suboptimal due to the off-policy and offline characteristics.

\section{DistillDirect}
\label{sec:DistillDirect}
\subsection{Comparison of Online vs. Offline and On-Policy vs. Off-Policy Training} 
When creating a preference dataset $\sD$, for any given prompt $\vx$, initially, two responses $\vy^1$ and $\vy^2$ are generated from an LLM denoted as $\genparam$. These responses are then assessed for preference by humans or AI, being labeled as $\vy^+$  (preferred) and $\vy^-$ (rejected). In this context, training is considered \textit{on-policy} if $\genparam$ = $\policyparam$, or when the generated responses are sampled from the latest version of the LLM during RLAIF training. \textit{Off-policy} training indicates otherwise. Learning is deemed \textit{online} if the preference labeling and training is conducted in real-time, directly on the outputs from the currently trained policy. It is considered \textit{offline} if preference labeling and training are performed in separate, discrete steps. 

 A critical caveat of employing a pre-collected preference dataset for offline and off-policy training, such as Distilled DPO, is distribution shifts \cite{guo2024direct}. More specifically, distribution shifts arise at time step $t$ because the preferred and rejected response is sampled from a policy $\genparam$, where $\genparam \neq \policyparamt$. Research has shown that online RLAIF systems, which are designed to mitigate these distribution shifts, significantly surpass the effectiveness of offline DPO methodologies \cite{guo2024direct}.

\subsection{RLAIF with DistillDirect}
\begin{figure}[!htb]
    \captionsetup{labelfont=bf={bf}}
    \centering
    \includegraphics[width=\columnwidth]{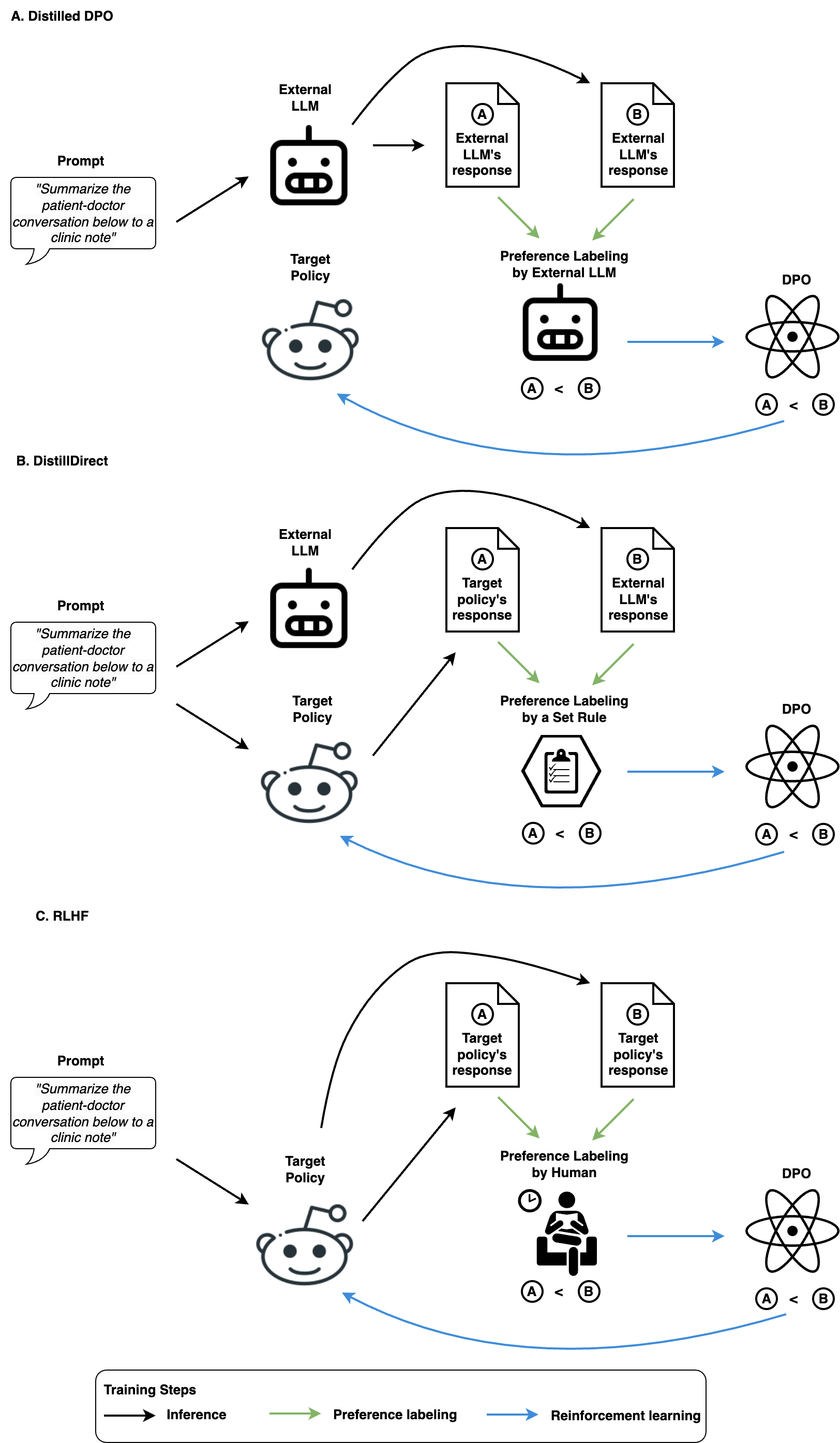}
    \caption{\textbf{Comparison of Distilled DPO, DistillDirect, and RLHF.} \textbf{A. Distilled DPO:} Preference dataset is generated and labeled by external LLMs rather than by the target policy, resulting in off-policy and offline training.
    \textbf{B. DistillDirect:} A response is generated from the target policy for each prompt, thereby making training on-policy. Additionally, another response is generated from an external LLM serving as the teacher model. \textbf{C. RLHF:} All responses are generated by the target policy, and preference labeling is completed by humans. Consequently, the training process is on-policy and online. In our study, we utilized DistillDirect for on-policy learning of RLAIF followed by further online and on-policy learning using RLHF.}
    
    \label{fig:distilldirect}
\end{figure}

We introduce an improved approach based on distilled DPO, termed DistillDirect, specifically designed to ensure on-policy learning on a distilled dataset. This approach is inspired by recent developments in adversarial preference optimization \cite{cheng2023adversarial}, online AI feedback \cite{guo2024direct}, iterative DPO \cite{pang2024iterative, xiong2024iterative} and SPIN \cite{chen2024self}. In each training cycle, we begin by sampling a response from the current policy $\policyparamt$, ensuring that the learning process remains strictly on-policy. This sampled response is then designated as the rejected response, while a reference response from Gemini Pro (the teacher model) is considered the preferred outcome (Figure \ref{fig:distilldirect}B). This approach implicitly assumes that the response from $\policyparamt$ is generally less favored than that from the teacher model—an assumption that we validated through manual review in each round of training. 

In our study, we employed RLAIF with DistillDirect, followed by Reinforcement Learning from Human Feedback (RLHF), which provides several advantages. First, for each iteration of DistillDirect, we used the target policy's outputs as ``reject samples'' to inform feedback, ensuring on-policy learning as opposed to the off-policy training in previous work \cite{tunstall2023zephyr}. Second, during the RLHF phase, we gathered human preferences on responses generated by the target policy, promoting online and on-policy training  (Figure \ref{fig:distilldirect}C). 

\section{Experiments}\label{experiments}
\subsection{Experimental Design}
\noindent{\bf Model Selection:} We selected Meta's LLaMA-2-13B and conducted experiments using both the base and chat models \cite{touvron2023llama}. Gemini 1.0 Pro (hereafter referred to as Gemini Pro) by Google was selected as the teacher model for generating reference notes \cite{team2023gemini}.

\noindent{\bf Experiments Overview:}
We demonstrated the experiments pipeline in Figure \ref{fig:overview}. First, we undertook domain-specific adaptation of LLaMA-2-13B through continued pretraining on discharge summaries from MIMIC-IV. This was followed by task-specific fine-tuning with SFT through instruction tuning. Next, we conducted RLAIF using DistillDirect, our enhanced approach to performing DPO on a distilled dataset. We then selected the model that performed best after SFT and RLAIF, as measured by ROUGE scores against reference notes, for RLHF via DPO. Finally, a panel of physicians conducted a blinded evaluation of the notes authored by our LLaMA-Clinic model, Gemini Pro, and other physicians. All training was performed using low-rank adaptation (LoRA).

\subsection{Dataset and Preprocessing}
\noindent{\bf Modified ACI-BENCH:} Given above mentioned limitation of ACI-BENCH, we established a simple yet specific note format, recognized as a ``best practice'' by a panel of licensed internal medicine physicians, to standardize our training approach. For model training, we retained only the dialogue section from ACI-BENCH and employed Gemini Pro to generate notes based on the ``best practice'' format, serving as our reference notes. We demonstrated two examples of clinical notes before and after the change in Appendix Figure \ref{fig:app.note_style}.

\noindent{\bf Dialogue-G:} We created a synthetic dataset of clinical dialogue-note pairs using Gemini Pro. This dataset, named {\it Dialogue-G}, comprises 1,291 cases. We first compiled transcribed outpatient notes from the publicly available synthetic MTSamples dataset \cite{MTSamples, hu2024improving} and utilized Gemini Pro to transform these notes into dialogues. Subsequently, we used these dialogues as inputs for Gemini Pro once again, this time to generate clinical notes based on the ``best practice'' format. 

\noindent{\bf MIMIC-IV:} MIMIC-IV encompasses 431,231 unique hospital admissions from Beth Israel Deaconess Medical Center in Boston, Massachusetts \cite{johnson2023mimic}. We utilized discharge summaries from MIMIC-IV for continued pretraining. Notably, the ``brief hospital course'' section of the discharge summaries is structurally akin to the ``assessment and plan'' section in outpatient notes. We compiled a subset of discharge summaries with only the ``brief hospital course'' using methods detailed in \cite{wang2024drg}, referred to as {\it Discharge-short}. We denoted the complete discharge summaries dataset as {\it Discharge-long}.

\subsection{Experiment setup}
\noindent{\bf Continued pretraining:} We explored both the Discharge-long dataset (1.2 billion tokens) and the Discharge-short dataset (0.2 billion tokens). 

\noindent{\bf Dataset Split for SFT, RLAIF and RLHF:} We combined the training subsets from ACI-BENCH (dialogue n = 67) and Dialogue-G (dialogue n = 1291), then split this data equally for SFT and RLAIF, stratified by the data source. For RLHF, we utilized dialogues from the training, test2, and test3 subsets of ACI-BENCH (dialogue n = 147). 

\noindent{\bf Physician Preference Data Collection:} 
In each round of RLHF, for a specific prompt $\vx$, three responses are generated from $\policyparamt$ and evaluated by our physician reviewers. Three licensed internal medicine physicians are tasked with providing preference feedback by selecting the most and least preferred responses, with criteria focusing on clinical readiness, correctness, and adherence to the desired format. A notable adaptation in our approach is that reviewers are also instructed to make adjustments to improve the quality of the preferred responses, such as correcting factual inaccuracies. 

\noindent{\bf Physician Reader Study:} The three internal medicine physicians engaged in preference data collection were tasked with writing clinical notes based on conversations from the ACI-BENCH test1 subset (dialogue n = 40), adhering to the pre-defined ``best practice'' format. The physician-authored notes, alongside those generated by Gemini Pro and LLaMA-Clinic, were reviewed by another four physicians, who were not involved in the preference labeling.

\section{Results}\label{results}

\subsection{Analysis of Continued Pretraining}
\begin{figure}[!h]
    \captionsetup{labelfont=bf={bf}}
    \centering
    \renewcommand{\figurename}{Figure}
    \includegraphics[width=\columnwidth]{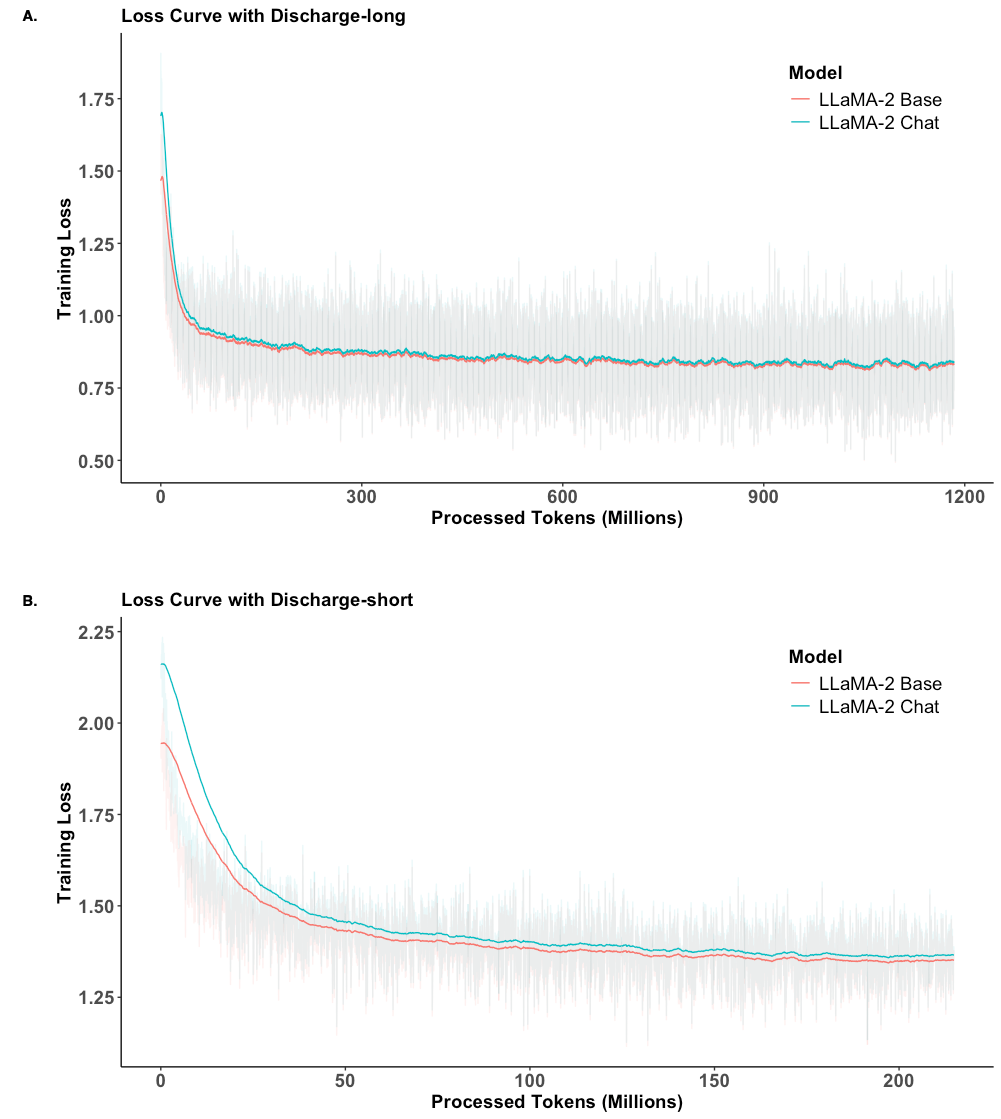}
    \caption{\textbf{Training Loss Curve from Continued Pretraining}. A. Training with the Discharge-long dataset (1.2 billion tokens). B. Training with the Discharge-short dataset (0.2 billion tokens). The X-axis represents processed training tokens, and the Y-axis represents training loss. The figures illustrate results from mixed precision training with a cosine learning rate scheduler. All experiments were trained for 1 epoch on their respective training datasets. The loss curve in the solid line was smoothed with an exponential moving average and a window size of 250 steps. The original loss values are shown as the faded background.}
    \label{fig:loss_curve}
\end{figure}

We presented the training loss curve in Figure \ref{fig:loss_curve}. Across all experiments, the training loss rapidly decreased after the initial few hundred steps, then leveled off, showing minimal improvement thereafter. The lowest training loss achieved with the Discharge-long dataset is approximately 0.9, whereas, with the Discharge-short dataset, it remained around 1.4. The trajectories of training loss were similar for both the chat and base models. When experimenting with various training strategies and hyperparameters, we frequently observed loss spikes that were slow to recover, as shown in Appendix Figure \ref{fig:app.loss_spike}. We proceeded with checkpoints that did not exhibit loss spikes for SFT and RLAIF.

\subsection{Analysis of SFT and RLAIF}

\begin{table*}[ht]
\centering
\small
\renewcommand{\arraystretch}{1.1}
\captionsetup{
    labelfont={bf},
    textfont={}
}
\setlength{\tabcolsep}{3pt} 
\resizebox{\textwidth}{!}{
\begin{tabular}{l|cccc|cccc}
\toprule
& \multicolumn{4}{c|}{\textbf{Subjective}} & \multicolumn{4}{c}{\textbf{Assessment and Plan}} \\
\textbf{Model} & \textbf{ROUGE-1} & \textbf{ROUGE-2} & \textbf{ROUGE-L} & \textbf{ROUGE-LSUM} & \textbf{ROUGE-1} & \textbf{ROUGE-2} & \textbf{ROUGE-L} & \textbf{ROUGE-LSUM} \\
\midrule
\midrule
\multicolumn{8}{l}{\textbf{Baseline: Out-of-Box LLaMA-2 and Medical LLMs}} \\
\midrule
\textbf{13B} & 0.0329 & 0.0033 & 0.0211 & 0.0268 & 0.0100 & 0.0006 & 0.0058 & 0.0085 \\
\textbf{13B-chat} & 0.3585 & 0.1281 & 0.2103 & 0.2890 & 0.4543 & 0.1778 & 0.2898 & 0.4261 \\
\textbf{Meditron-7B} & 0.1249 & 0.0052 & 0.0678 & 0.0957 & 0.0895 & 0.0032 & 0.0527 & 0.0810 \\
\textbf{LLaMA3-Med42-8B} & 0.2496 & 0.0662 & 0.1392 & 0.2056 & 0.2372 & 0.0577 & 0.1426 & 0.2187 \\
\textbf{MeLLaMA-13B-chat} & 0.2756 & 0.1268 & 0.1925 & 0.2329 & 0.2655 & 0.0964 & 0.1865 & 0.2522 \\
\midrule
\multicolumn{8}{l}{\textbf{SFT + RLAIF Models (No Pretraining)}} \\
\midrule
\textbf{13B} & & & & & & & & \\
\hspace{10pt}SFT & 0.2813 & 0.1166 & 0.1975 & 0.2248 & 0.2977 & 0.1059 & 0.1963 & 0.2677 \\
\hspace{10pt}SFT + DistillDirect & \textbf{0.4994} & \textbf{0.2633} & \textbf{0.3425} & \textbf{0.3964} & 0.4941 & 0.2398 & 0.3476 & 0.4650 \\
\textbf{13B-chat} & & & & & & & & \\
\hspace{10pt}SFT & 0.2874 & 0.1179 & 0.2022 & 0.2285 & 0.3301 & 0.1191 & 0.2231 & 0.2992 \\
\hspace{10pt}SFT + DistillDirect & 0.4783 & 0.2472 & 0.3252 & 0.3738 & 0.4893 & \textbf{0.2411} & \textbf{0.3674} & 0.4599 \\
\midrule
\multicolumn{8}{l}{\textbf{Continued Pretraining + SFT + RLAIF Models}} \\
\midrule
\textbf{13B-long} & & & & & & & & \\
\hspace{10pt}CP & 0.1612 & 0.0139 & 0.0754 & 0.1378 & 0.1289 & 0.0108 & 0.0598 & 0.1178 \\
\hspace{10pt}CP + SFT & 0.2525 & 0.1034 & 0.1714 & 0.2044 & 0.2565 & 0.0775 & 0.1651 & 0.2276 \\
\hspace{10pt}CP + SFT + DistillDirect & 0.4494 & 0.2471 & 0.3224 & 0.3618 & 0.4578 & 0.2086 & 0.3272 & 0.4241 \\
\textbf{13B-short} & & & & & & & & \\
\hspace{10pt}CP & 0.1488 & 0.0106 & 0.0712 & 0.1149 & 0.0899 & 0.0060 & 0.0449 & 0.0806 \\
\hspace{10pt}CP + SFT & 0.2463 & 0.0858 & 0.1638 & 0.1951 & 0.2277 & 0.0619 & 0.1396 & 0.1960 \\
\hspace{10pt}CP + SFT + DistillDirect & 0.4775 & 0.2561 & \textbf{0.3452} & 0.3842 & \textbf{0.4956} & 0.2328 & 0.3526 & \textbf{0.4663} \\
\textbf{13B-chat-long} & & & & & & & & \\
\hspace{10pt}CP & 0.1708 & 0.0151 & 0.0792 & 0.1321 & 0.1536 & 0.0130 & 0.0746 & 0.1424 \\
\hspace{10pt}CP + SFT & 0.3463 & 0.1591 & 0.2525 & 0.2837 & 0.3620 & 0.1237 & 0.2401 & 0.3290 \\
\hspace{10pt}CP + SFT + DistillDirect & 0.4601 & 0.2504 & 0.3325 & 0.3773 & 0.4662 & 0.2395 & 0.3484 & 0.4498 \\
\textbf{13B-chat-short} & & & & & & & & \\
\hspace{10pt}CP & 0.1520 & 0.0116 & 0.0725 & 0.1085 & 0.1162 & 0.0107 & 0.0630 & 0.1032 \\
\hspace{10pt}CP + SFT & 0.3475 & 0.1310 & 0.2230 & 0.2711 & 0.3055 & 0.1036 & 0.2009 & 0.2824 \\
\hspace{10pt}CP + SFT + DistillDirect & \textbf{0.4878} & \textbf{0.2613} & 0.3410 & \textbf{0.3883} & \textbf{0.5182} & \textbf{0.2689} & \textbf{0.3933} & \textbf{0.4915} \\
\bottomrule
\end{tabular}
}
\caption{\textbf{ROUGE Scores Following Continued Pretraining (CP), Supervised Fine-Tuning (SFT), and Reinforcement Learning from AI Feedback (RLAIF).} All evaluations were conducted on the validation subset of the modified ACI-BENCH. The ``13B'' models represent the LLaMA-2 base models, while the ``13B-chat'' models correspond to the LLaMA-2 chat models. Models with the suffix ``long'' were pretrained using the Discharge-long dataset, and those with the suffix ``short'' were pretrained using the Discharge-short dataset. \textbf{Bolded} scores represent the top two scores for each metric. Continued pretraining with MIMIC-IV discharge summaries initially resulted in performance deterioration, whereas SFT and DistillDirect significantly improved the performance across all models.}
\label{table:s/p RLAIF}
\end{table*}

Since the primary objective of SFT and RLAIF is to align LLaMA-2's output with Gemini Pro, we evaluated ROUGE scores—a measure of lexical similarity—against reference notes generated by Gemini Pro. We reported ROUGE scores \cite{lin2004rouge} post-SFT and RLAIF, alongside those from baseline models in Table \ref{table:s/p RLAIF}. Interestingly, at baseline, medical LLMs fine-tuned on biomedical literature or clinical notes \cite{chen2023meditron, christophe2024med42, xie2024me} underperform compared to the vanilla LLaMA-2 chat model. This suggests that generating notes aligned with our ``best practice'' format presents an out-of-distribution challenge for medical LLMs. As expected, continued pretraining with MIMIC-IV discharge summaries compromised the chat model's capacity to follow instructions. SFT notably enhanced performance, particularly for the chat model compared to the base model. The application of RLAIF with DistillDirect significantly boosted performance across all models. In many cases, the gains achieved through RLAIF surpassed those from SFT, as reflected by the delta improvement in ROUGE scores (e.g., 13B-short). Notably, our training with DistillDirect frequently encountered instability—a well-known challenge in reinforcement learning \cite{ding2020challenges}. With carefully selected training hyperparameters, DistillDirect could achieve performance improvements without overfitting, despite the limited training data and repetitive use of the same prompts in each training cycle (see Appendix Table \ref{table:app.RLHF3_t1.0}). We have detailed our experiments, including ablation studies, to find a stable training setup in Appendix Method \ref{app.SFT_RLAIF_setup}. 

When comparing models pretrained with Discharge-long and Discharge-short, the latter consistently achieved higher ROUGE scores, despite exhibiting higher training loss during the pretraining phase. The chat model pretrained with the Discharge-short dataset (13B-chat-short) emerged as the top performer for the final RLHF stage. Interestingly, the model trained with SFT and RLAIF, without continued pretraining, demonstrated strong performance, including the highest score for ``Subjective.'' This raises questions about the benefits of continued pretraining, particularly considering its substantial computational demands (Appendix Figure \ref{fig:app.cost}A).

\subsection{Analysis of RLHF} 
We collected quantitative feedback from physician reviewers, who identified inaccurate information and hallucinations as the most common issues in the LLM-generated notes. Two out of three reviewers noted a significant improvement in the quality of the notes, with fewer hallucinations after one round of DPO. Consequently, we conducted only two rounds of DPO, mindful of its time-intensive nature (Appendix Figure \ref{fig:app.cost}A). We present a qualitative analysis of a specific case across different stages of model training, as shown in Appendix Figure \ref{fig:app.note_progression}. This example illustrates that RLHF introduced more granular changes, building on the outcomes of RLAIF while preserving the original style and structure. We named the model after our RLHF step as LLaMA-Clinic. 

\subsection{Analysis of Physician Reader Study}

\begin{figure*}[!ht]
    \captionsetup{labelfont=bf={bf}}
    \centering
    \includegraphics[width=\textwidth]{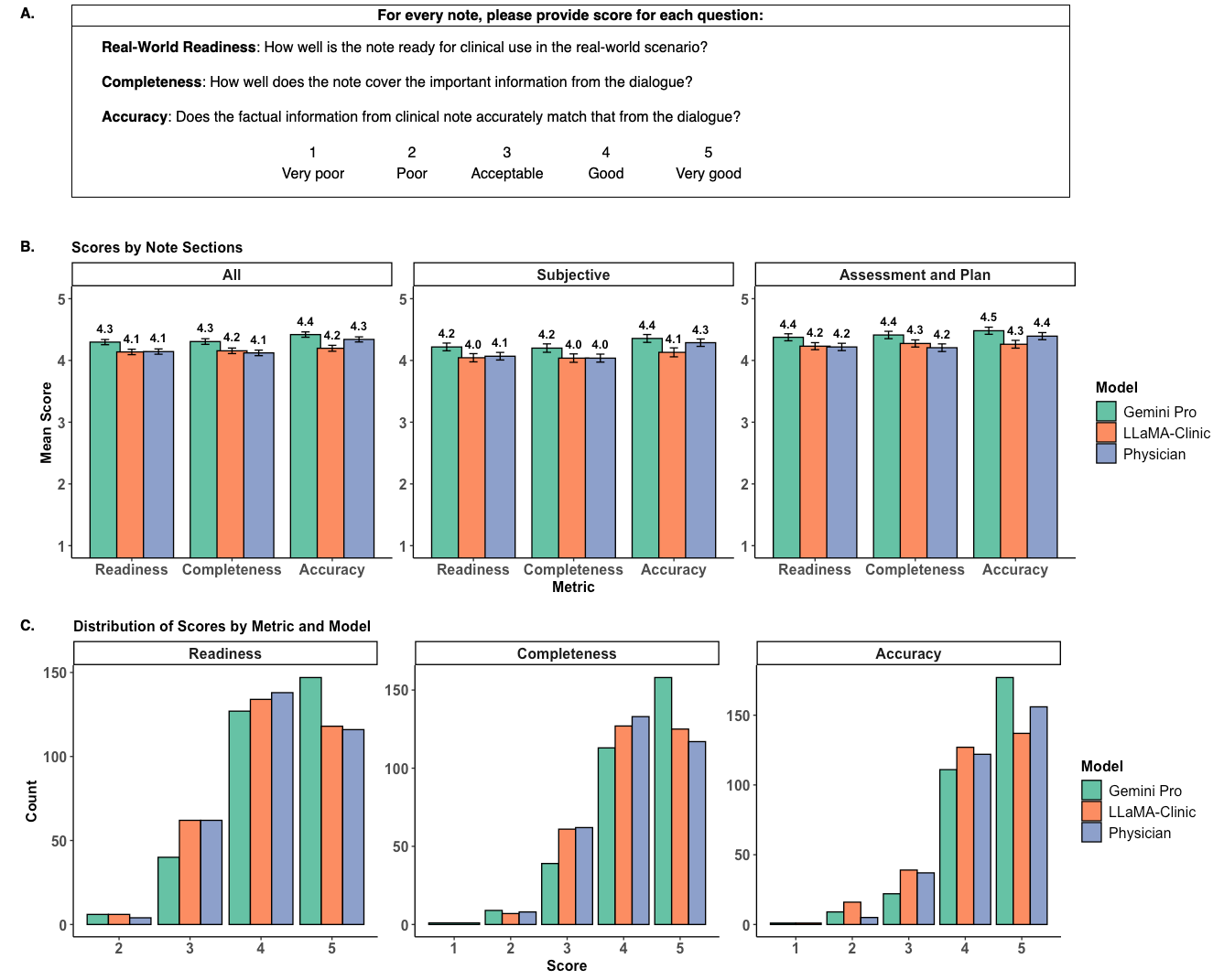}
    \caption{\textbf{Physician Reader Study}. A. Survey questions assessing each clinical note on three criteria: real-world readiness, completeness, and accuracy. B. Bar charts displaying the mean score among evaluators across different sections of the note. The error bars represent standard errors. The X-axis represents various metrics, and the Y-axis shows the mean scores. The subfigure labeled ``All'' displays the combined results. C. Bar charts displaying the distribution of scores across all criteria. The X-axis represents different scores, and the Y-axis shows total counts.}
    
    \label{fig:main_result}
\end{figure*}

We presented the results from the physician reader study in Figure \ref{fig:main_result}. Gemini Pro achieved the highest scores across all three criteria (Figure \ref{fig:main_result}B). Remarkably, the majority (92.8\%) of the individual evaluations rated the notes generated by LLaMA-Clinic as ``acceptable'' or higher across all three criteria (Figure \ref{fig:main_result}C). Furthermore, the overall distribution of scores was also similar among the three groups. This aligns with qualitative feedback from three out of four reviewers, who stated that the quality of notes was, for the majority of the time, indistinguishable among the groups, particularly in the ``Subjective'' section.

The metric of real-world readiness yielded intriguing observations. Physicians were asked to evaluate the notes as though they would be utilized in everyday clinical practice, assuming the physicians would proofread and make necessary edits. LLaMA-Clinic achieved comparable overall readiness scores to physician-authored notes, including in the more complex ``Assessment and Plan'' section, as shown in Figure~\ref{fig:main_result}B. Overall, LLaMA-Clinic received higher completeness scores but lower accuracy scores compared to physician-authored notes. This finding suggests that physicians might be more tolerant of minor factual inaccuracies in a real-world setting if the notes require fewer edits.

\subsection{Cost Analysis for Model Development and Inference}
As shown in Appendix Figure \ref{fig:app.cost}A, the majority of compute hours for developing LLaMA-Clinic were required during the continued pretraining stage. However, unsurprisingly, the most time-consuming step overall involved physician labeling during RLHF. Based on pricing information from May 2024, LLaMA-Clinic showed a 3.75-fold reduction in inference costs compared to its teacher model, Gemini 1.0 Pro (Appendix Figure \ref{fig:app.cost}B).


\section{Discussion}\label{discussion}
Our methodology for adapting LLMs to clinical note generation revealed several key considerations. We initiated this process by establishing a ``best practice'' note format informed by a consensus among our physician authors. This approach tackles the variability in the style, format, and quality of physician notes, which could otherwise compromise the training of LLMs. Instead of relying on the LLM to identify the most effective note structure, we advocate for healthcare providers to set these standards themselves. It is important to acknowledge that the ``best practice'' note format employed in this study reflects the consensus of a group of internal medicine physicians within a single organization. This format may not be generalizable to other specialties or other organizations, such as orthopedic surgery. Furthermore, we examined a workflow in which providers critically review and refine AI-generated notes prior to their filing. Under such a workflow, physician preferences may shift toward notes requiring fewer revisions and edits, with a higher tolerance for minor inaccuracies. 
 
\section{Conclusions}
Our research highlights the potential of training an open-source LLM for outpatient note generation, showing strong promise for real-world clinical applications. Healthcare institutions are in a privileged position to undertake such endeavors, given their access to extensive EMR data and a wealth of domain expertise critical for implementing RLHF. Our work was based on fewer than 1,500 patient-doctor dialogues and limited physician preference data. When implementing a similar project in a healthcare institution, the training data could be scaled up significantly, potentially leading to further performance improvements. Lastly, the prospect of extending this work to other clinical note-generation tasks, such as creating discharge summaries for hospitalized patients, is particularly exciting.

\section{Acknowledgments}
This research was supported by NSF awards SCH-2205289. The funder played no role in the study design, data collection, analysis, and interpretation of data, or the writing of this manuscript. This research is also supported by the Protected Research Time Grant from the Southwest Minnesota Region of the Mayo Clinic Health System.

\section{Ethical Considerations}
We followed the same approach as in \cite{van2023clinical, walsh2017measuring} to assess the potential consequences of factual errors. One physician evaluated the likelihood and severity of harm associated with notes that received an accuracy or completeness score below 5 in his review (n = 22, 31 and 39 for Gemini Pro, LLaMA-Clinic, and Physician, respectively). All notes within the LLaMA-clinic and physician groups were assigned scores of ``None'' for the extent of harm and ``Low'' for the likelihood of harm. In contrast, the Gemini Pro group contained a single case rated with ``Mild to Moderate'' extent of harm and a ``Medium'' likelihood of harm.

\section{Limitations}
Our study serves as a proof of concept and encounters limitations, notably the scarcity of publicly available patient-doctor dialogues for model training, with the largest ACI-BENCH dataset comprising fewer than 300 cases. Additionally, the effectiveness of outpatient note generation is fundamentally linked to the content of patient-doctor dialogues. Although the ACI-BENCH data may have synthesized dialogues to encapsulate all necessary information for composing a comprehensive note, such ideal conditions may not always reflect real-world scenarios. For instance, time constraints may prevent physicians from discussing every detail of medical reasoning with patients, potentially degrading the quality of the generated notes due to the suboptimal input dialogue. Our research was also constrained by limited availability of physician evaluators, which restricted our ability to conduct extensive hyperparameter searches or additional rounds of RLHF. Our final evaluation was conducted with only four physicians due to the task's time-intensive nature.


\bibliography{main}

\appendix
\setcounter{table}{0}
\setcounter{figure}{0}

\clearpage

\section{Experiment Chronicles}
\subsection{Overview}
Our objective is to perform domain- and task-specific adaptation of LLaMA-2 for generating outpatient clinical notes based on patient-physician dialogues. This process is structured around four sequential steps: continued pretraining (CP), supervised finetuning (SFT), reinforcement learning from AI feedback (RLAIF), and reinforcement learning from human feedback (RLHF). Here, we detailed the trial-and-error process throughout our experiments. 

\subsection{Continued Pretaining (CP)}
\subsubsection{CP Experiment 1}
\noindent{\bf Models:} LLaMA-2-13B base and chat models with Low-Rank Adaptation (LoRA).

\noindent{\bf Dataset:} We conducted experiments on two datasets: the complete discharge summary from MIMIC-IV (Discharge-long) and the extracted ``Hospital Course'' section from discharge summaries (Discharge-short).

\noindent{\bf Approach:} In this experiment, we evaluated the efficacy of pure bf16 training while adhering to the majority of the default hyperparameters specified in the LLaMA-recipes. We used a learning rate (LR) of 3e-4 without a scheduler. Other hyperparameters were selected based on computing resource available at the time of experiments (4 x A6000 or 4 x A100 GPUs), including a global batch size of 16 without gradient accumulation. 

\begin{figure}[!ht]
    \captionsetup{labelfont=bf={bf}}
    \centering
    \renewcommand{\figurename}{Appendix Figure}
    \includegraphics[width=\columnwidth]{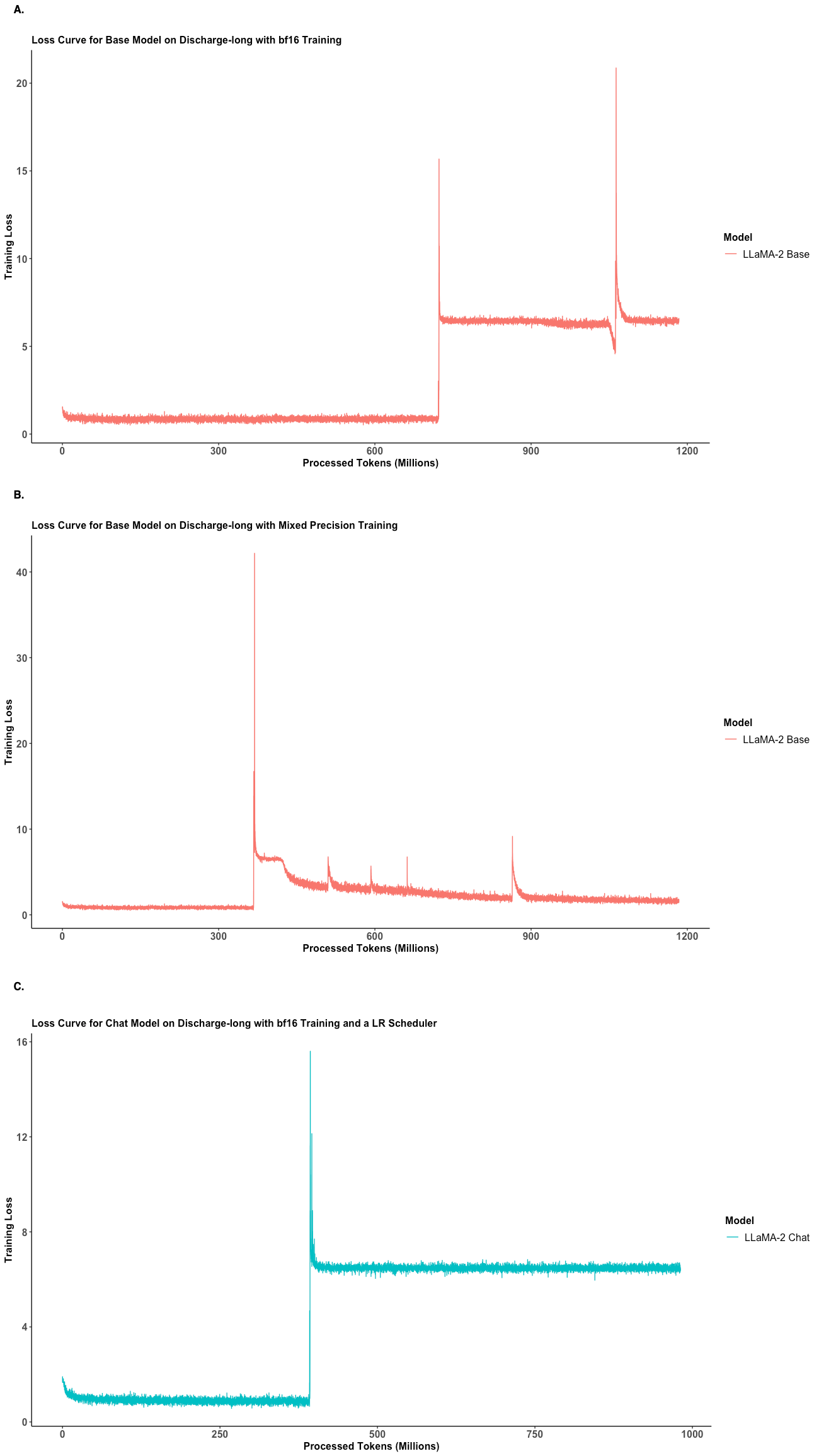}
    \caption{\textbf{Training Loss Spikes for 13B Models during Continued Pretraining}. A. Loss curve for base model with bf16 training on vanilla LLaMA-recipes without a LR scheduler. B. Loss curve for base model with mixed precision training without a LR scheduler. C. Loss curve for chat model with a LR scheduler and bf16 training. All experiments were performed on Discharge-long dataset. The X-axis represents processed training tokens, and the Y-axis represents training loss. Original loss curves were shown without smoothing.}
    \label{fig:app.loss_spike}
\end{figure}

\noindent{\bf Results:} We ran into training loss spike on 13B base model that never fully recovered (Appendix Figure \ref{fig:app.loss_spike} A). 

\noindent{\bf Solution:} Restarted pretraining using a mixed precision strategy.

\subsubsection{CP Experiment 2}
\noindent{\bf Models and Dataset:} Same as experiment 1. 

\noindent{\bf Approach:} In this experiment, we implemented mixed precision training, wherein weights and gradients were stored in bf16 format, and optimizer states in fp32. This decision was based on the hypothesis that training with pure bf16 might pose challenges in achieving convergence \cite{zamirai2020revisiting}. The remaining setup parameters were consistent with those outlined in Experiment 1.

\noindent{\bf Results:} During the course of training the 13B-base model, we encountered another spike in training loss, notably in the early stages of the training process (Appendix Figure \ref{fig:app.loss_spike}B). Although the loss quick recovered, it did not revert to its baseline level. For runs that were completed successfully, the outcomes were comparable to those observed in Experiment 1.

\noindent{\bf Solution:} Restarted pretraining using a LR scheduler.

\subsubsection{CP Experiment 3}
\noindent{\bf Models and Dataset:} Same as experiment 1. 

\noindent{\bf Approach:} The standard LLaMA-recipes library does not incorporate a LR scheduler at the time of our work, resulting in a constant LR for each epoch in the previous two experiments. In this iteration, we maintained mixed precision training and introduced a cosine LR scheduler with 200 warm-up steps, followed by a decay to 0\% of the peak LR.

\noindent{\bf Results:} All runs were completed successfully, with the loss curve presented in Figure \ref{fig:loss_curve}. We observed a significant reduction in training loss when using the Discharge-long dataset, with losses dropping to the 0.9 range. Conversely, for the Discharge-short dataset, the training loss plateaued around 1.4. \textbf{We utilized checkpoints from this experiment for SFT and RLAIF}.

\subsubsection{CP Experiment 4}
\noindent{\bf Models and Dataset:} Same as experiment 1. 

\noindent{\bf Approach:} In this experiment, we explored the implementation of pure bf16 training, utilizing the same cosine scheduler as in Experiment 3. Our aim was to ascertain whether this approach could stabilize pure bf16 training, which offers the advantage of reduced VRAM requirements.

\noindent{\bf Results:} A spike in training loss was observed in the 13B-chat model during training with the Discharge-Long dataset (Appendix Figure \ref{fig:app.loss_spike}C).

\subsubsection{CP Experiment 5}
\noindent{\bf Models:} LLaMA-2-7B base and chat models with full parameter training.  

\noindent{\bf Dataset:} Same as experiment 1. 

\noindent{\bf Approach:} Motivated by the inferior performance of LoRA compared to full parameter training across various benchmarks \cite{han2023medalpaca}, we explored the implementation of full parameter training on 7B models, considering their similar VRAM requirements (compared to LoRA training on the 13B models).

\noindent{\bf Results:} In the 7B-chat model runs, we observed a spike in training loss for both mixed precision and pure bf16 training modalities, despite employing a cosine LR scheduler.

\noindent{\bf Solution:} Lowered LR. 

\subsubsection{CP Experiment 6}
\noindent{\bf Models:} Same as experiment 5. 

\noindent{\bf Dataset:} Same as experiment 1. 

\noindent{\bf Approach:} We reduced LR to 2e-5, down from 3e-4 in the previous experiment. Of note, the vanilla LLaMA-2 7B model's used a LR of 3e-4 for pretraining \cite{touvron2023llama}.

\noindent{\bf Results:} With the adjusted LR, along with the implementation of pure BF16 training and a cosine LR scheduler, we were able to successfully complete all runs. Checkpoints from these runs were utilized for SFT and RLAIF.

\subsection{SFT and RLAIF} \label{app.SFT_RLAIF_setup}
\subsubsection{SFT Experiment 1}
\noindent{\bf Models:} We utilized the four 13-B checkpoints (base-model/chat-model trained with Discharge-long/Discharge-short) from CP Experiment 3 employing LoRA, and the four 7-B checkpoints from CP Experiment 6 with full parameter training.

\noindent{\bf Dataset:} We combined the training subsets from ACI-BENCH (dialogue n = 67) and Dialogue-G (dialogue n = 1291), then split this data equally for SFT and RLAIF, stratified by data source. Notably, for each dialogue we ask model to generate ``Subjective'' and ``Assessment and Plan'' in two separate prompts (therefore two data points per dialogue).

\noindent{\bf Approach:} Instruction tuning was applied to the SFT dataset over 3 epochs, with a LR of 2e-5 for all models, maintaining a fixed LR.

\noindent{\bf Results:} The performance metrics post-SFT are presented in Appendix Table \ref{table:app.RLHF1}, under the column labeled ``R0''. Interestingly, the models, after undergoing pretraining and SFT, exhibited inferior performance compared to the out-of-the-box LLaMA-2-chat models. This decline in performance may be attributed to catastrophic forgetting observed during the CP phase.

\begin{table*}[!ht]
\small
\renewcommand{\arraystretch}{1.2}
\setlength{\tabcolsep}{7pt}
\renewcommand{\tablename}{Appendix Table}
\captionsetup{labelfont=bf={bf}}
\resizebox{\linewidth}{!}{
\begin{tabular}{lllll|llll}
\toprule 
 & \multicolumn{4}{c}{\textbf{Subjective}} & \multicolumn{4}{c}{\textbf{Assessment and Plan}} \\
Models & R0 & R1 & R2 & R3 & R0 & R1 & R2 & R3 \\
\hline
\hline
\multicolumn{9}{l}{\textit{Models underwent continued pretraining, SFT and RLAIF}} \\
\quad 7B-short & 0.2332 & 0.3373 & 0.3163 & 0.2730 & 0.2259 & 0.2209 & 0.3164 & 0.3410 \\
\quad 7B-long & 0.2308 & 0.3619 & 0.0875 & 0.3262 & 0.2030 & 0.1992 & 0.2357 & 0.3205 \\
\quad 7B-chat-short & 0.2644 & 0.3471 & 0.0292 & \textbf{0.4465} & 0.2531 & 0.2670 & 0.3370 & \textbf{0.4436} \\
\quad 7B-chat-long & 0.2322 & 0.4270 & 0.4072 & 0.2639 & 0.2305 & 0.2841 & 0.3193 & 0.3916 \\
\quad 13B-short & 0.1335 & 0.3472 & 0.2353 & 0.3757 & 0.1586 & 0.1873 & 0.0540 & 0.3079 \\
\quad 13B-long & 0.1929 & 0.3329 & 0.2234 & 0.2949 & 0.1816 & 0.2602 & 0.2367 & 0.2918 \\
\quad 13B-chat-short & 0.2161 & 0.3703 & 0.4168 & 0.2297 & 0.2411 & 0.3901 & 0.3629 & 0.3202 \\
\quad 13B-chat-long & 0.1797 & 0.3617 & 0.3342 & 0.4065 & 0.2337 & 0.3381 & 0.3944 & 0.3903 \\
\hline
\multicolumn{9}{l}{\textit{Out-of-box models}} \\
\quad 7B\ & \multicolumn{4}{c}{0.0305} & \multicolumn{4}{c}{0.000} \\
\quad 7B-chat & \multicolumn{4}{c}{0.2808} & \multicolumn{4}{c}{0.3538} \\
\quad 13B\ & \multicolumn{4}{c}{0.0249} & \multicolumn{4}{c}{0.0049} \\
\quad 13B-chat & \multicolumn{4}{c}{0.3114} & \multicolumn{4}{c}{0.3693} \\
\bottomrule
\end{tabular}
}
\\
\caption{\textbf{ROUGE-1 Score after RLAIF Experiment 1.} Performance reported on validation subset of original ACI-BENCH. R0 represent the model after SFT, with or without CP. R1 to R3 represents performance after respective rounds of DistillDirect. Models ended in long were pretrained using Discharge-long dataset, while models ended in short were pretrained using Discharge-short dataset. \textbf{Bolded scores} denote the best performance with respect to the task. The experimental setup includes training on the original reference notes from ACI-BENCH with a LR of 2e-5, and 3 epochs per training round. The temperature is set at 1.0 during generation time to calculate ROUGE-1.}
\label{table:app.RLHF1}
\end{table*}

\subsubsection{RLAIF Experiment 1}
\noindent{\bf Models:} We continued with the eight checkpoints derived from SFT Experiment 1. For the 13B models, LoRA was employed, while full parameter training was applied to the 7B models.

\noindent{\bf Dataset:} The RLAIF split of the training dataset as mentioned above.

\noindent{\bf Approach:} RLAIF was conducted using DistillDirect over three rounds. In each round, we first generated notes using our training models, which were designated as ``rejected'' notes. For ACI-BENCH, the ``preferred notes'' were the original reference notes in ACI-BENCH. For Dialogue-G, the preference notes were generated by Gemini-pro. For subsequent rounds, we sampled from the newly updated model checkpoint to produce a new set of ``rejected'' notes, while maintaining the same ``preferred'' notes. Each round consisted of three epochs, with a LR of 2e-5 in addition to a cosine scheduler.

\noindent{\bf Results:} Performance metrics following RLAIF Experiment 1 are presented in Appendix Table \ref{table:app.RLHF1}. Although we achieved commendable performance by the third round of DistillDirect (notably the best performance came from the 7B-chat\_short model), the training process exhibited instability. For instance, the performance in the second round of DistillDirect for the 7B-chat-short model showed a significant decline from the previous round. Across all models, we did not observe a consistent enhancement in performance (as assessed by the ROUGE-1 score on the validation set) with additional rounds of DPO training. Notably, training accuracy reached 100\% prematurely, during the first epoch for all models with a high reward margin (as shown in Appendix Figure \ref{fig:app.DPO}), indicating potential overfitting.

\noindent{\bf Solution:} Recognizing the instability of the current DistillDirect setup, we considered several potential causes, including issues with data quality/distribution shift, an inappropriate LR leading to overfitting, and reward hacking, among others. Our initial step towards addressing these concerns involved a thorough examination of the training data.

\begin{figure}[t]
\includegraphics[width=\columnwidth]{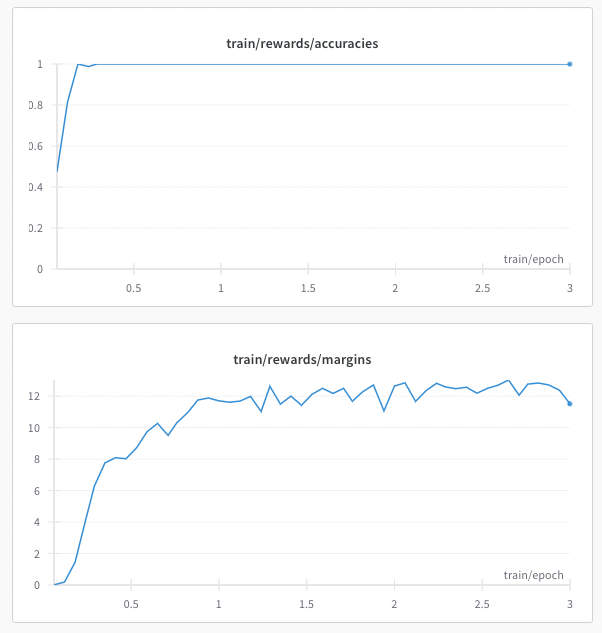}
\captionsetup{labelfont=bf={bf}}
\centering
\renewcommand{\figurename}{Appendix Figure}
\caption{\textbf{Example Training Set Accuracy and Reward Margin during DPO with a LR of 2e-5.} Examples taken from 13B-chat-short\_R3, and all other runs have similar training curve with high accuracy and reward margin early on with this LR.}
\label{fig:app.DPO}
\end{figure}

\subsubsection{SFT Experiment 2}
\noindent{\bf Models:} We restarted SFT using the eight checkpoints after CP as discussed in SFT experiment 1.

\noindent{\bf Dataset:} Upon closely examining the ACI-BENCH and our synthetic dataset, we observed several key points: 
\begin{enumerate}
      \item The quality of reference notes within ACI-BENCH is markedly variable. Certain notes are of poor quality (for example, overly brief), and others adopt syntax or styles not typically found in real clinical notes (for instance, explicitly including a ``medical reasoning'' section).
      \item Additionally, notes from different subspecialties exhibit significant variations in style and format (e.g., a note from internal medicine versus one from orthopedics). While the heterogeneity observed in ACI-BENCH likely mirrors the diversity encountered in clinical practice (where different physicians may write notes very differently), it poses a challenge to use these notes as a ``gold standard'' for training a language model to replicate such documentation.
      \item Similar issues were also noted in our augmented Dialogue-G dataset created using Gemini-pro. Despite employing prompts with clear instructions regarding note format, minor inconsistencies still emerged.
  \end{enumerate}

\noindent{\bf Approach:} Through discussions with our physician co-authors, we concluded that before AI can assist physicians in generating notes, the physicians themselves must determine what constitutes a ``best practice'' note format. It's acknowledged that the ``best practice'' might diverge from existing notes for valid reasons (notably, recognizing that physicians can also produce suboptimal notes). With this perspective, we revised our prompts and provided a single-shot example to Gemini-pro to foster ``constrained'' note generation tailored towards a specific style. This new note style was collectively endorsed by our physician authors as superior to many examples within ACI-BENCH. Consequently, we directed Gemini-pro to generate new reference notes from the dialogues in ACI-BENCH, resulting in a dataset we named Modified ACI-BENCH. At this stage, we also regenerated reference notes for Dialogue-G using Gemini Pro. Initially, we considered having our physicians edit Gemini-pro's notes before employing them for training. However, we abandoned this idea due to the time-intensive nature of the task and the satisfactorily high quality of outputs from Gemini-pro with our revised prompt.

\noindent{\bf Results:} Utilizing the newly formulated modified ACI-BENCH dataset, we performed SFT for 3 epochs, adhering to the same parameters as before. The performance metrics post-SFT are detailed in Appendix Table \ref{Table:RLHF2}, under the column labeled ``R0''.

\begin{table*}[!ht]
\small
\renewcommand{\arraystretch}{1.2}
\setlength{\tabcolsep}{7pt}
\renewcommand{\tablename}{Appendix Table}
\captionsetup{labelfont=bf={bf}}
\resizebox{\linewidth}{!}{
\begin{tabular}{p{2.5cm}llll|llll}
\toprule 
 & \multicolumn{4}{c}{\textbf{Subjective}} & \multicolumn{4}{c}{\textbf{Assessment and Plan}} \\
 Models & R0 & R1 & R2 & R3 & R0 & R1 & R2 & R3 \\
 \hline
 \hline
\multicolumn{9}{l}{\textit{Models underwent continued pretraining, SFT and RLAIF}} \\
\quad 7B-short & 0.1998 & 0.4894 & 0.5198 & 0.3792 & 0.2617 & 0.4885 & 0.0917 & 0.4100 \\
\quad 7B-long & 0.1949 & 0.4688 & 0.0594 & 0.3920 & 0.2446 & 0.3705 & 0.3053 & 0.4040 \\
\quad 7B-chat-short & 0.2961 & 0.4816 & 0.4946 & 0.0586 & 0.3157 & 0.5201 & 0.4763 & 0.3867 \\
\quad 7B-chat-long & 0.3067 & 0.5051 & 0.0964 & 0.0773 & 0.2988 & 0.4907 & 0.4275 & 0.3628 \\
\quad 13B-short & 0.2463 & 0.5064 & 0.4904 & 0.4066 & 0.2277 & 0.5252 & 0.4059 & 0.4493 \\
\quad 13B-long & 0.2525 & 0.4199 & 0.5113 & 0.3309 & 0.2565 & 0.4840 & 0.4610 & 0.3566 \\
\quad 13B-chat-short & 0.3475 & 0.5115 & 0.5030 & \textbf{0.5441} & 0.3055 & 0.5277 & 0.4422 & \textbf{0.5422} \\
\quad 13B-chat-long & 0.3463 & 0.4856 & 0.5308 & 0.3863 & 0.3620 & 0.4752 & 0.4730 & 0.4633 \\
\hline
\multicolumn{9}{l}{\textit{Models underwent SFT and RLAIF without continued pretraining}} \\
\quad 13B & 0.2813 & 0.5303 & 0.3967 & 0.5108 & 0.2977 & 0.5265 & 0.3510 & 0.5060 \\
\quad 13B-chat & 0.2874 & 0.5073 & 0.5287 & \textbf{0.5452} & 0.3301 & 0.5069 & 0.4895 & \textbf{0.5597} \\
\hline
\multicolumn{9}{l}{\textit{Out-of-box models}} \\
\quad 13B\ & \multicolumn{4}{c}{0.0329} & \multicolumn{4}{c}{0.0100} \\
\quad 13B-chat & \multicolumn{4}{c}{0.3585} & \multicolumn{4}{c}{0.4543} \\
\bottomrule
\end{tabular}
}
\\
\caption{\textbf{ROUGE-1 Score after RLAIF Experiment 2.} Performance reported on validation subset of modified ACI-BENCH. R0 represent the model after SFT, with or without CP. R1 to R3 represents performance after respective rounds of DistillDirect. Models ended in long were pretrained using Discharge-long dataset, while models ended in short were pretrained using Discharge-short dataset. \textbf{Bolded scores} denote the best performance with respect to the task. The experimental setup includes training on the new reference notes (modified ACI-BENCH) with a LR of 2e-5, and 3 epochs per training round.  The temperature is set at 1.0 during generation time to calculate ROUGE-1. }
\label{Table:RLHF2}
\end{table*}

\subsubsection{RLAIF Experiment 2}
\noindent{\bf Models:} The eight checkpoints from SFT experiment 2. We used LoRA for 13B models and full parameter training for 7B models. 

\noindent{\bf Dataset:} New dataset as described in SFT experiment 2.

\noindent{\bf Approach:} Similar to RLAIF Experiment 1, but this iteration utilized the newly generated notes by Gemini-pro as ``preferred'' notes. Additionally, to mitigate concerns of overfitting, we limited the training to 1 epoch for each round, as opposed to the 3 epochs per round in RLAIF Experiment 1.

\noindent{\bf Results:} Several intriguing observations emerge from this iteration (Appendix Table \ref{Table:RLHF2}). 
  \begin{itemize}
      \item \textbf{7B vs 13B models:} With the current LR, the 7B models exhibited significant instability during DistillDirect. For instance, the 7B-short model demonstrated a notable performance degradation from round 1 to round 2. Upon manual evaluation of its outputs, the model incorrectly generated ``Subjective'' sections when tasked with producing ``Assessment and Plan'' and frequently returned null outputs. Given the more stable training observed with 13B models utilizing LoRA, along with their superior performance and reduced storage requirements compared to full parameter training for 7B models, \textbf{we have decided to proceed exclusively with 13B models henceforth}.
      \item \textbf{Best Performer by ROUGE-1 :} Among our CP models, the 13B-chat model pre-trained with the Discharge-short dataset emerged as the top performer. However, the overall best model was the 13B-chat model following SFT and three rounds of DistillDirect, \textbf{but without CP}. It achieved ROUGE-1 scores of 0.5452 for ``Subjective'' and 0.5597 for ``Assessment and Plan''. Naturally, this led us to question the value of CP in our specific task. Such skepticism seemed justified, particularly given the differences between our training corpus of discharge summaries and that of outpatient clinical notes, which could be deemed out-of-distribution for our task.
      \item \textbf{Possible Best Performer by Physician Review:} The quality of the notes generated by the top-performing models (e.g., 13B-chat\_R3, 13B-short\_R1, 13B-chat-short\_R3, 13B-chat-long\_R2) appears nearly indistinguishable to our physician author. Overall, we find the ``Subjective'' sections of these notes to be satisfactory and potentially suitable for clinical application. However, the primary deficiency in achieving clinical-ready notes lies within the ``Assessment and Plan'' sections, particularly regarding medical reasoning and certain linguistic nuances. Although somewhat subjective, we felt the outputs from 13B-short\_r1 bear the greatest resemblance to an actual provider's note. \textbf{This finding underscores the limitations of quantitative metrics in tasks involving domain-specific language generation}.
      \item \textbf{Persistent DistillDirect Training Instability:} Despite the creation of a more uniform dataset, DistillDirect training remained susceptible to fluctuations. This instability was more pronounced for 7-billion parameter models but was also observed in 13-billion parameter models (Appendix Table \ref{Table:RLHF2}). The 13B-long model serves as an example, where the ROGUE-1 score dropped from 0.5113 in round 2 to 0.3309 in round 3 for the ``Subjective'' section. Reducing the number of epochs per DistillDirect round from three to one did not mitigate this issue. In all DistillDirect rounds, training set accuracy rapidly reached a perfect score of 1.0 halfway through the first epoch, accompanied by a high reward margin (as illustrated in the initial third of the curve shown in Appendix Figure \ref{fig:app.DPO}). This observation again suggests potential overfitting. Interestingly, similar rapid accuracy increases were reported during Zephyr training, and overfitting did not negatively impact performance on downstream tasks \cite{tunstall2023zephyr}.
      \item \textbf{Reflection on LR Selection:} The optimal LR for DPO has yet to be established in literature, with only a limited number of LLMs having undergone DPO training to date. Notably, given the small size of our training set, our initial LR choice of 2e-5 is larger than those selected in Zephyr and Tulu-2, where both models opted for an LR of 5e-7 \cite{ivison2023camels, tunstall2023zephyr}. The Tulu-2 paper mentioned that a ``slow LR... is required for stable and effective DPO training.'' Conversely, in another study, Gaudi-2 employed a significantly higher LR (5e-4, compared to an LR of 1e-4 during its SFT), despite using the same training data as Zephyr, and still achieved a favorable response \cite{intel_analytics_2023}. 
    \item \textbf{Generative Parameter Matters:} We investigated the influence of generative parameters on model performance.  Specifically, we experimented with a lower temperature setting of 0.6 compared to the baseline of 1.0 (Appendix Table \ref{Table:app.RLHF2_t0.6}), while maintaining other parameters the same (multinomial sampling with top-k = 50, top-p = 1.0, and repetition penalty = 1.2). Lowering the temperature resulted in a consistent increase in ROUGE-1 scores across all models. This finding suggests that optimizing generation-related hyperparameters through a dedicated search process has the potential to further enhance performance.      
  \end{itemize}

\noindent{\bf Solution:} Conduct a limited LR search specifically for DistillDirect.

\begin{table*}[!ht]
\small
\renewcommand{\arraystretch}{1.2}
\setlength{\tabcolsep}{7pt}
\renewcommand{\tablename}{Appendix Table}
\captionsetup{labelfont=bf={bf}}
\resizebox{\linewidth}{!}{
\begin{tabular}{p{2.5cm}llll|llll}
\toprule 
 & \multicolumn{4}{c}{\textbf{Subjective}} & \multicolumn{4}{c}{\textbf{Assessment and Plan}} \\
 Models & R0 & R1 & R2 & R3 & R0 & R1 & R2 & R3 \\
 \hline
 \hline
\multicolumn{9}{l}{\textit{Models underwent continued pretraining, SFT and RLAIF}} \\
\quad 13B-short & 0.4006 & 0.5423 & 0.5186 & 0.4702 & 0.4032 & 0.5683 & 0.4039 & 0.5242 \\
\quad 13B-long & 0.3570 & 0.4799 & \textbf{0.5779} & 0.3981 & 0.4298 & 0.5266 & 0.4688 & 0.3780 \\
\quad 13B-chat-short & 0.4390 & 0.5455 & 0.5248 & 0.5560 & 0.4825 & 0.5565 & 0.5236 & \textbf{0.5694} \\
\quad 13B-chat-long & 0.4272 & 0.5149 & 0.5423 & 0.4188 & 0.4459 & 0.5487 & 0.5322 & 0.4852 \\
\hline
\multicolumn{9}{l}{\textit{Models underwent SFT and RLAIF without continued pretraining}} \\
\quad 13B & 0.3671 & 0.5469 & 0.4216 & 0.5508 & 0.4335 & 0.5804 & 0.3833 & 0.4913 \\
\quad 13B-chat & 0.4077 & 0.5351 & \textbf{0.5582} & 0.5466 & 0.4383 & 0.5522 & 0.4907 & \textbf{0.5847} \\
\bottomrule
\end{tabular}
}
\\
\caption{\textbf{ROUGE-1 Score after RLAIF Experiment 2 with Lower Temperature.} Performance reported on validation subset of modified ACI-BENCH. R0 represent the model after SFT, with or without CP. R1 to R3 represents performance after respective rounds of DistillDirect. Models ended in long were pretrained using Discharge-long dataset, while models ended in short were pretrained using Discharge-short dataset. \textbf{Bolded scores} denote the best performance with respect to the task. The experimental setup includes training on the new reference notes (modified ACI-BENCH) with a LR of 2e-5, and 1 epoch per training round. The temperature is set at 0.6 during generation time to calculate ROUGE-1, using the same model checkpoints as in Appendix Table \ref{Table:RLHF2}.}
\label{Table:app.RLHF2_t0.6}
\end{table*}

\subsubsection{RLAIF Experiment 3}
\noindent{\bf Models:} 13B model checkpoints from SFT experiment 2.

\noindent{\bf Dataset:} New dataset as described in SFT experiment 2.

\noindent{\bf Approach:} We evaluated an LR of 5e-6 and 5e-7 and compared the effects of 1 epoch versus 3 epochs in each round across three rounds. After each round of training, we resampled outputs from the updated model checkpoint to serve as ``rejected'' samples for the subsequent round of DPO, while continuing to use notes produced by Gemini-pro as the ``preferred'' samples.

\noindent{\bf Results:} \begin{itemize}
      \item \textbf{LR of 5e-7:} Training set accuracy reached 0.98 by the end of the second epoch. After three epochs, the reward margin only attained a level of 0.25 (Appendix Figure \ref{fig:app.DPO-newLR}). This very small, albeit popular, LR was found to be insufficient in our experimental setup, yielding very slow performance improvements regardless of whether we conducted 1 epoch per round or 3 epochs per round of training (Appendix Table \ref{table:app.RLHF3_t1.0}).

\begin{figure}[h]
\includegraphics[width=\columnwidth]{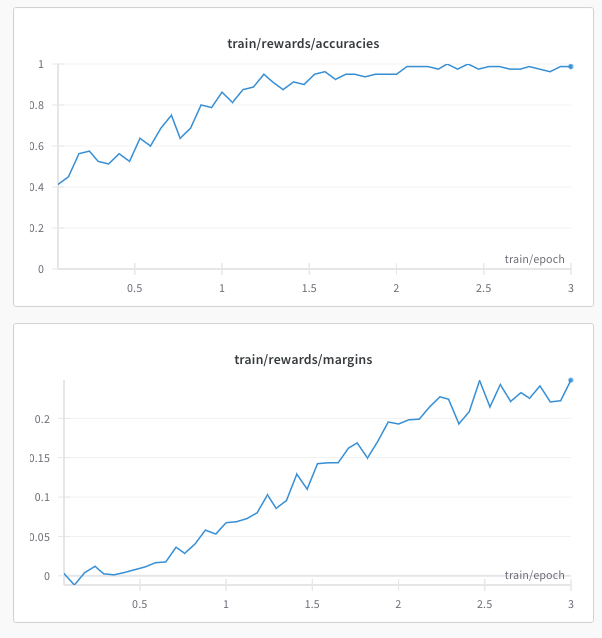}
\centering
\captionsetup{labelfont=bf={bf}}
\renewcommand{\figurename}{Appendix Figure}
\caption{\textbf{Example Training Set Accuracy and Reward Margin during DPO with a LR of 5e-7.} Examples taken from 13B-chat-short\_R1. Training set accuracy reaches 0.98 at the end of 2nd epoch. After 3 epochs, the reward margin only reaches a level of 0.25.}
\label{fig:app.DPO-newLR}
\end{figure}

\begin{figure}[!h]
\includegraphics[width=\columnwidth]{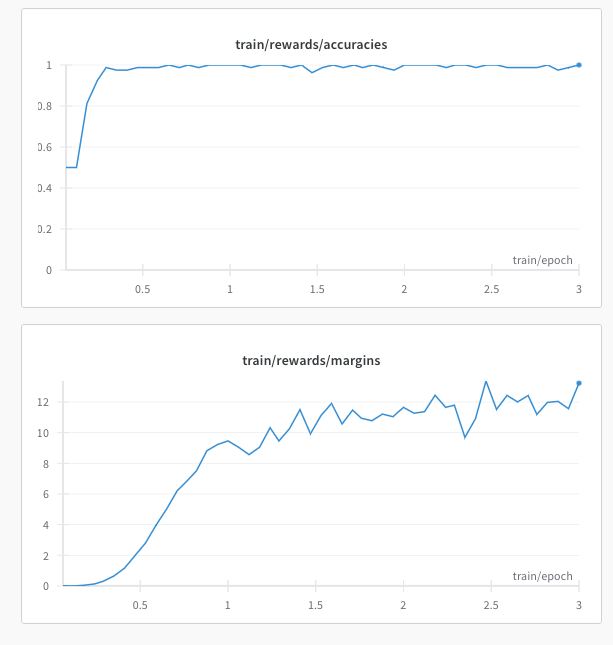}
\centering
\captionsetup{labelfont=bf={bf}}
\renewcommand{\figurename}{Appendix Figure}
\caption{\textbf{Example Training Set Accuracy and Reward Margin During DPO with a LR of 5e-6.} Examples taken from 13B-chat-short\_R1. Training set accuracy reaches 0.98 at the end of 2nd epoch. After 3 epochs, the reward margin only reaches a level of 0.25.}
\label{fig:app.DPO-midLR}
\end{figure}
      
      \item \textbf{LR of 5e-6:} Training set accuracy reached 0.95 midway through the second epoch. After three epochs, the reward margin achieved a level of 12 (refer to Appendix Figure \ref{fig:app.DPO-midLR}). Although the training curve bears resemblance to that associated with a LR of 2e-5 (see Appendix Figure \ref{fig:app.DPO-midLR} and Appendix Figure \ref{fig:app.DPO}), this lower LR demonstrated greater stability. This was particularly evident in the experiments conducted with 1 epoch per round of training, where performance across all models either exhibited steady improvement or achieved convergence, devoid of the abrupt declines observed with an LR of 2e-5 (Appendix Table \ref{table:app.RLHF3_t1.0}). Notably, several models reached their performance peak in round 2, indicating that 2 to 3 rounds of training might be an optimal range for our experiment setup.
      \item \textbf{Impact of Temperature:} Consistent with previous experiments, model performance improves when a lower temperature was applied during generation (see Appendix Table \ref{table:app.RLHF3_t0.6}). 
  \end{itemize} 
  \noindent{\bf Solution:} We identified the LR of 5e-6 with a 1-epoch per round as the stable setup for DistillDirect in our experiment. Upon manual examination of outputs from all model checkpoints within this configuration, \textbf{we selected the 13B-chat-short model from round 3 for the next stage of RLHF} (Appendix Table \ref{table:app.RLHF3_t1.0}).

\begin{table*}[!ht]
\small
\renewcommand{\arraystretch}{1.2}
\setlength{\tabcolsep}{7pt}
\renewcommand{\tablename}{Appendix Table}
\captionsetup{labelfont=bf={bf}}
\resizebox{\linewidth}{!}{
\begin{tabular}{p{2.5cm}llll|llll}
\toprule 
 & \multicolumn{4}{c}{\textbf{Subjective}} & \multicolumn{4}{c}{\textbf{Assessment and Plan}} \\
 Models & R0 & R1 & R2 & R3 & R0 & R1 & R2 & R3 \\
 \hline
 \hline
 \multicolumn{9}{l}{\textit{Lr: 5e-6, 3 epochs per round}} \\
\quad 13B-short & 0.2463 & 0.5124 & 0.4820 & 0.5310 & 0.2277 & 0.5123 & 0.4547 & 0.4808 \\
\quad 13B-long & 0.2525 & 0.4406 & \textbf{0.5423} & 0.2974 & 0.2565 & 0.4652 & 0.4984 & 0.3316 \\
\quad 13B-chat-short & 0.3475 & 0.4633 & 0.4727 & 0.5422 & 0.3055 & \textbf{0.5293} & 0.5012 & 0.5086 \\
\quad 13B-chat-long & 0.3463 & 0.4936 & 0.5097 & 0.5327 & 0.3620 & 0.5290 & 0.5008 & 0.5247 \\
 \multicolumn{9}{l}{\textit{Lr: 5e-6, 1 epoch per round}} \\
 \hline
\quad 13B-short & 0.2463 & 0.3708 & 0.4794 & 0.4775 & 0.2277 & 0.4318 & 0.4995 & 0.4956 \\
\quad 13B-long & 0.2525 & 0.3730 & 0.4427 & 0.4494 & 0.2565 & 0.3609 & 0.4750 & 0.4578 \\
\quad 13B-chat-short & 0.3475 & 0.4254 & 0.4804 & \textbf{0.4878} & 0.3055 & 0.4329 & 0.5140 & \textbf{0.5182} \\
\quad 13B-chat-long & 0.3463 & 0.4135 & 0.4600 & 0.4601 & 0.3620 & 0.4642 & 0.4843 & 0.4662 \\
\hline
 \multicolumn{9}{l}{\textit{Lr: 5e-7, 3 epochs per round}} \\
\quad 13B-short & 0.2463 & 0.2546 & 0.2655 & 0.2744 & 0.2277 & 0.2618 & 0.2573 & 0.3002 \\
\quad 13B-long & 0.2525 & 0.2151 & 0.2382 & 0.2115 & 0.2565 & 0.2458 & 0.3149 & 0.2755 \\
\quad 13B-chat-short & 0.3475 & 0.3518 & \textbf{0.3718} & 0.3428 & 0.3055 & 0.3398 & 0.3449 & \textbf{0.3840} \\
\quad 13B-chat-long & 0.3463 & 0.3311 & 0.2984 & 0.3296 & 0.3620 & 0.3310 & 0.3719 & 0.3747 \\
\hline
 \multicolumn{9}{l}{\textit{Lr: 5e-7, 1 epoch per round}} \\
\quad 13B-short & 0.2463 & 0.2864 & 0.2921 & 0.2644 & 0.2277 & 0.2599 & 0.2731 & 0.2735 \\
\quad 13B-long & 0.2525 & 0.2600 & 0.2333 & 0.1839 & 0.2565 & 0.2473 & 0.2216 & 0.2453 \\
\quad 13B-chat-short & \textbf{0.3475} & 0.3117 & 0.3388 & 0.3356 & 0.3055 & 0.3303 & 0.3370 & 0.3231 \\
\quad 13B-chat-long & 0.3463 & 0.3090 & 0.2943 & 0.3134 & 0.3620 & 0.3174 & 0.3194 & \textbf{0.3693} \\
\bottomrule
\end{tabular}
}
\\
\caption{\textbf{ROUGE-1 Score after RLAIF Experiment 3.} Performance reported on validation subset of modified ACI-BENCH. R0 represent the model after CP and SFT. DistillDirect R1 to R3 represents performance after respective rounds of DistillDirect. Models ended in long were pretrained using Discharge-long dataset, while models ended in short were pretrained using Discharge-short dataset. \textbf{Bolded scores} denote the best performance with respect to the task. The experimental setup includes training on the new reference notes (modified ACI-BENCH) with a variable learning rates and training epochs. The temperature is set at 1.0 during generation time to calculate ROUGE-1.}
\label{table:app.RLHF3_t1.0}
\end{table*}

\subsection{RLHF}
The model checkpoints selected for RLHF have undergone training through a sequence of stages: CP Experiment 3, followed by SFT Experiment 2, and then RLAIF Experiment 3. During the RLHF phase, we experimented with different temperature settings and observed that physicians tend to prefer outputs generated at lower temperatures. Consequently, we decided to adopt more deterministic generation parameters for the physician reader study.

\begin{table*}[!ht]
\small
\renewcommand{\arraystretch}{1.2}
\setlength{\tabcolsep}{7pt}
\renewcommand{\tablename}{Appendix Table}
\captionsetup{labelfont=bf={bf}}
\resizebox{\linewidth}{!}{
\begin{tabular}{p{2.5cm}llll|llll}
\toprule
 & \multicolumn{4}{c}{\textbf{Subjective}} & \multicolumn{4}{c}{\textbf{Assessment and Plan}} \\
 Models & R0 & R1 & R2 & R3 & R0 & R1 & R2 & R3 \\
 \hline
 \hline
 \multicolumn{9}{l}{\textit{Lr: 5e-6, 3 epochs per round}} \\
\quad 13B-short & 0.4006 & 0.5165 & 0.5143 & 0.5349 & 0.4032 & 0.5471 & 0.4996 & 0.5246 \\
\quad 13B-long & 0.3570 & 0.5049 & 0.5679 & 0.4289 & 0.4298 & 0.5064 & 0.5102 & 0.3976 \\
\quad 13B-chat-short & 0.4390 & 0.5343 & 0.5309 & 0.5614 & 0.4825 & 0.5524 & 0.5266 & \textbf{0.5575} \\
\quad 13B-chat-long & 0.4272 & 0.4815 & \textbf{0.5759} & 0.5605 & 0.4459 & 0.5299 & 0.5510 & 0.5294 \\
\hline
 \multicolumn{9}{l}{\textit{Lr: 5e-6, 1 epoch per round}} \\
\quad 13B-short & 0.4006 & 0.4971 & 0.5145 & 0.5200 & 0.4032 & 0.4950 & 0.5464 & 0.5222 \\
\quad 13B-long & 0.3570 & 0.4449 & 0.5126 & 0.5000 & 0.4298 & 0.5123 & 0.5190 & 0.5285 \\
\quad 13B-chat-short & 0.4390 & 0.5155 & 0.5167 & \textbf{0.5352} & 0.4825 & 0.5380 & 0.5392 & \textbf{0.5411} \\
\quad 13B-chat-long & 0.4272 & 0.4863 & 0.4980 & 0.5077 & 0.4459 & 0.5094 & 0.5255 & 0.5346 \\
\hline
 \multicolumn{9}{l}{\textit{Lr: 5e-7, 3 epochs per round}} \\
\quad 13B-short & 0.4006 & 0.3760 & 0.3771 & \textbf{0.4573} & 0.4032 & 0.4390 & 0.4281 & 0.4315 \\
\quad 13B-long & 0.3570 & 0.4022 & 0.3909 & 0.3693 & 0.4298 & 0.4154 & 0.4428 & 0.4411 \\
\quad 13B-chat-short & 0.4390 & 0.4434 & 0.4482 & 0.4394 & 0.4825 & 0.4761 & 0.4625 & \textbf{0.4833} \\
\quad 13B-chat-long & 0.4272 & 0.4398 & 0.4058 & 0.4282 & 0.4459 & 0.4763 & 0.4633 & 0.4545 \\
\hline
 \multicolumn{9}{l}{\textit{Lr: 5e-7, 1 epoch per round}} \\
\quad 13B-short & 0.4006 & 0.4241 & 0.4112 & 0.4140 & 0.4032 & 0.4345 & 0.4172 & 0.4107 \\
\quad 13B-long & 0.3570 & 0.4099 & 0.4175 & 0.4039 & 0.4298 & 0.3671 & 0.3865 & 0.4285 \\
\quad 13B-chat-short & 0.4390 & 0.4291 & \textbf{0.4541} & 0.4337 & 0.4825 & 0.4569 & \textbf{0.4826} & 0.4450 \\
\quad 13B-chat-long & 0.4272 & 0.4188 & 0.4048 & 0.3986 & 0.4459 & 0.4606 & 0.4552 & 0.4446 \\
\bottomrule
\end{tabular}
}
\\
\caption{\textbf{ROUGE-1 Score after RLAIF Experiment 3 with a Lower Temperature.} Performance reported on validation subset of modified ACI-BENCH. R0 represent the model after CP and SFT. R1 to R3 represents performance after respective rounds of DistillDirect. Models ended in long were pretrained using Discharge-long dataset, while models ended in short were pretrained using Discharge-short dataset. \textbf{Bolded scores} denote the best performance with respect to the task. The experimental setup includes training on the new reference notes (modified ACI-BENCH) with a variable learning rates and training epochs. The temperature is set at 0.6 during generation time to calculate ROUGE-1 using the same model checkpoints as in Appendix Table \ref{table:app.RLHF3_t1.0}.}
\label{table:app.RLHF3_t0.6}
\end{table*}

\clearpage
\clearpage
\section{Details on Datasets}
\subsection{ACI-BENCH Subsets}
ACI-BENCH comprises five data subsets: train, validation, test1, test2 and test3 \cite{yim2023aci}. Test1 and test2 corresponds to the test sets from ACL ClinicalNLP MEDIQA-Chat 2023 TaskB and TaskC, respectively \cite{abacha2023overview}. Test3 corresponds to TaskC of CLEF MEDIQA-SUM 2023 \cite{yim2023overview}. Given the scarcity of publicly available clinical dialogue-note datasets, we used the train, test2, and test3 subsets for various stages of model development in our study. The blinded clinical reader study was performed on the test1 subset. 

\subsection{MIMIC-IV}
The publicly available MIMIC-IV dataset comprises 431,231 unique hospital admissions from 299,712 patients admitted to an ICU or the ED of the Beth Israel Deaconess Medical Center in Boston, Massachusetts \cite{johnson2023mimic}. MIMIC-IV is deidentified according to the Health Insurance Portability and Accountability Act (HIPAA) Safe Harbor provision \cite{johnson2023mimic}. Access to MIMIC-IV can be requested at \url{https://physionet.org/content/mimiciv/},
which requires a signed safe usage agreement.

\section{Implementation Details for Final Experimental Steps}
\subsection{LoRA}
We used LoRA to train LLaMA-2-13B models for all phases of training. LoRA is a method that involves freezing the pre-trained model weights and only training a small percentage (<1\%) of the model weights, i.e., by incorporating trainable rank decomposition matrices into each layer of the transformer architecture \cite{hu2021lora}. 
As a quick summary, let us assume that we have the original weight matrix $\bm{W_0} \in \mathbb{R}^{d \times k}$. LoRA works by adding a low-rank matrix to the original weight matrix: $\Delta \bm{W} + \bm{W_0}, \Delta \bm{W}  = \bm{B}\bm{A}$ where $\bm{B}\in \mathbb{R}^{d \times r}$ and $\bm{A}\in \mathbb{R}^{r\times k}$. $r << d$, so the matrices $\bm{B}, \bm{A}$ are limited by a lower rank $r$, reducing the need to train all the parameters. Training is only performed on this $\Delta \bm{W}$, and original model weights are kept the same. We then scale $\Delta \bm{W}$ by $\frac{\alpha}{r}$, where $\alpha$ is a constant in $r$.

In all training steps, LoRA parameters were configured with $r$ set to 8, an $\alpha$ of 32, and a dropout rate of 0.05. All attention blocks were included in the LoRA target modules.

\subsection{Medical LLMs Evaluation} \label{app:medical LLMs}
We evaluated state-of-the-art medical and clinical LLMs fine-tuned on biomedical literature and clinical notes: Meditron-7B \cite{chen2023meditron}, LLaMA-3-Med42-8B \cite{christophe2024med42}, and MeLLaMA-13B-chat \cite{xie2024me}. The same generation configurations with one-shot prompting from our main experiments were applied, except for MeLLaMA, for which we used its default generation settings to achieve better performance.

\subsection{Continued Pretraining}
We followed the training scripts outlined in Meta's official LLaMA recipe repository \cite{LlamaRecipes}. We employed mixed-precision training with a batching strategy of packing and a context length of 4096 tokens. We utilized Fully Sharded Data Parallel (FSDP) on either 4 Nvidia A6000 or 4 Nvidia A100 GPUs. We maintained a batch size of 4 during training with a gradient accumulation step of 1. Consistent with LLaMA-2, we set a peak learning rate of 3e-4 for the continued pretraining stage. The AdamW optimizer with a cosine learning rate scheduler was used, and the model was trained for one epoch. The exponential moving average of training loss as shown in Figure \ref{fig:loss_curve} was calculated using the \texttt{pracma} package from R with a window size of 250. 

\subsection{SFT}
We used a similar experiment setup as continued pretraining, including following LLaMA-recipes to perform mixed precision training on 4 GPUs using FSDP~\cite{zhao2023pytorch}. For SFT, we selected the batching strategy of padding and trained on 3 epochs. Consistent with LLaMA-2, we set a peak learning rate of 2e-5.  We truncate prompt (including instruction and dialogue) at a max length of 3000 tokens, and truncate note to 1000 tokens. We set a value of -100 for labels on prompt tokens to zero out losses from prompts.

\subsection{RLAIF}
We utilized the \texttt{trl} library from Huggingface to conduct DistillDirect \cite{TRL2020}. Due to computational limitations, experiments were conducted on a single Nvidia A100 GPU with 80GB of graphics memory. To optimize memory usage, pure BF16 training was utilized with a micro-batch size of 1 and gradient accumulation steps of 8. Following a limited learning rate search detailed in Appendix \ref{app.SFT_RLAIF_setup} and Appendix Table \ref{table:app.RLHF3_t1.0}, a learning rate of 5e-6 was chosen. The optimizer used was \texttt{paged\_adamw\_32bit}. Within the \texttt{DPOTrainer} class, we set the beta hyperparameter to 0.1 and passed None to \texttt{ref\_model}. Three rounds of DistillDirect were performed, with each round involving one epoch of training.

Text generation was implemented using the Transformers library \cite{wolf2020huggingfaces}. When generating ``rejected'' samples from the latest model checkpoint, we consistently applied do\_sample=True, top\_p=1.0, temperature=1.0, top\_k=50, repetition\_penalty=1.2, and use\_fast\_kernels=False. The maximum number of newly generated tokens was set to 1000.

\subsection{RLHF}
For RLHF, we employed an experimental setup analogous to that described in RLAIF. We conducted two rounds of DPO on human preference data. Diverging from the approach taken in RLAIF, we executed three epochs of training in each DPO round due to the limited size of the dataset.

Based on the findings from ablation studies detailed in Appendix \ref{app.SFT_RLAIF_setup} and Appendix Table \ref{table:app.RLHF3_t0.6}, we opted for a lower temperature setting at this stage. In the initial round of DPO, we generated three responses using the same configuration, including a temperature setting of 0.6, for preference labeling. In the second round of DPO, we diversified the temperature settings, resulting in three responses with temperatures set at 0.6, 0.4, and 0.2, respectively, for preference labeling.

Of note, we excluded data from Dialogue-G to ensure in-distribution training during this final stage of model development. This decision was based on the observed perplexity of 2.79 for Dialogue-G, in contrast to 5.62 for ACI-BENCH, as calculated using the LLaMA-2 chat model after continued pretraining.

\subsection{Physician Reader Study}
LLM-generated notes were produced by LLaMA-Clinic and Gemini Pro, employing identical generation-related hyperparameters (temperature of 0.2, top\_p of 0.7 and top\_k of 40). To ensure consistency in presentation across all notes, we implemented a basic post-processing step. This step standardized aspects like style, spacing, and capitalization to minimize any formatting discrepancies between human-authored and model-generated notes.

Four licensed physician evaluators, who specialize in general internal medicine or family medicine, boast rich experience in outpatient practice. The notes were presented in a random order, anonymized to remove any identifying information, and labeled as note 1, note 2, and note 3 to mask the origin of each note from the evaluators. Before assessing the notes, evaluators were instructed to read the entire patient-provider conversation. They were then asked to rate the quality of each note across three criteria: ``accuracy,'' ``completeness,'' and ``real-world readiness.'' For each criterion, a scoring system from 1 to 5 was used, ranging from very poor to very good, with higher scores reflecting superior quality. Specifically for ``real-world readiness,'' evaluators were prompted to consider the scenario of integrating AI-generated clinical notes into their daily practice, including the necessity to proofread and potentially edit these notes before filing them into medical records.

\subsection{Statistical Analysis}
The non-parametric Kruskal-Wallis H test was selected to compare differences in word counts among the three-note groups, utilizing the \texttt{scipy} package in Python \cite{virtanen2020scipy}. We measured IRR using Gwet's AC2 statistics implemented through the \texttt{irrCAC} package in R \cite{gwet2014handbook, gwet2019package}. We reported results with quadratic weights for Gwet's AC2, as this approach is reliable for ordinal data against the grey zones \cite{tran2021impact}. Due to our relatively small reviewer pool, we chose not to conduct statistical significance testing on the physician reader study, aligning with practices observed in the deep learning community \cite{touvron2023llama, stiennon2020learning}.

\subsection{Model Development Cost Estimation}
In Appendix Figure \ref{fig:app.cost}A, we provide cost estimations for the training steps directly involved in the development of LLaMA-Clinic. These costs should be viewed as minimal estimates and will likely fall short of the actual budget requirements since they do not include the trial-and-error expenses from various experiments, such as trialing different models, conducting hyperparameter searches, and debugging. The hours for continued pretraining are based on training using the Discharge-short dataset. For GPU hours, we accounted for the total number of Nvidia A100 GPU hours utilized. For example, if the continued pretraining stage requires 12 hours using FSDP on four A100 GPUs, we calculate this step as requiring 48 hours. For physician labeling hours, we asked physicians to estimate the average time they spent on the tasks.

\subsection{Model Inference Cost Estimation}
A complex array of factors influences the total cost of deploying a model for production. These include hardware and software configurations, labor costs associated with constructing, validating, and refining the model, and the implementation of comprehensive security measures to mitigate misuse and enhance threat detection. To ensure a fair and apples-to-apples comparison, we calculated the inference costs for both proprietary and open-source models based on API calls. In this context, the total annual inference cost is calculated as follows:
\begin{equation}
C = \left( p_i \times n_i + p_o \times n_o \right) \times R
\end{equation}

Here, $C$ represents the total annual inference cost. $p_i$ denotes the price per input token, while $n_i$ refers to the average number of input tokens per request. Similarly, $p_o$ indicates the price per output token, and $n_o$ represents the average number of output tokens per request. The term $R$ stands for the total number of annual requests. 

For open-source models, we assumed deployment on Fireworks.ai, a company that offers serverless inference for customized LLMs. We sourced pricing information from the websites of Google AI, OpenAI, and Fireworks.ai in May 2024 for on-demand API calls. Detailed pricing information can be found in Appendix Table \ref{tab:token_costs}. We assumed an average of 3,000 input tokens and 1,000 output tokens per request for clinical note generation. This estimation likely contains redundancy and leaves room for prompt engineering, given that in a cohort of real-world family medicine clinical encounters, the average lengths per dialogue and note are 1505 and 683 tokens, respectively \cite{yim2023aci}.

An important consideration for production is ensuring adequate throughput for LLMs. As an example, Gemini 1.0 Pro has a rate limit of 360 requests per minute, while LLaMA-Clinic, deployed with the ``Developer'' plan from Fireworks.ai, allows 600 requests per minute. We consider this default rate limit acceptable for our calculations, as shown in Figure 5. For example, 1 million annual requests translate into an average of approximately 5.7 requests per minute (RPM), using the formula:
\begin{equation}
\text{Average RPM} = \frac{\text{Annual Requests}}{365 \times 8 \times 60}
\end{equation}
assuming an 8-hour workday. However, this calculation does not account for peak demand, which would necessitate system redundancy in a production environment. In addition, there are other technical factors to consider for deployment in production, such as latency and throughput variance \cite{ArtificialAnalysisAI2024}, which were not included in our analysis.

\section{Additional Results}
\subsection{Performance of LLM-as-a-Judge in Predicting Physician Preference} \label{app:llm-as-a-judge}
We evaluated Gemini 1.0 as an LLM-as-a-judge on our RLHF dataset. We observed a low accuracy of 43.8\% in predicting physician preferences (from three candidate notes), underscoring the challenges LLMs encounter in predicting expert preferences in complex domains such as medicine.

\subsection{Qualitative Analysis of Model Outputs}
Qualitative analysis for a specific case at different stages of model training is presented in Appendix Figure \ref{fig:app.note_progression}. Continued pretraining effectively adopted the style and peculiarities from discharge summaries but at the expense of diminished instruction-following ability and increased hallucinations. The quality of outputs significantly improved post-SFT but remained overly verbose, while RLAIF effectively refined outputs to adhere to the format of reference notes, assisting in reducing hallucinations. Upon manually reviewing outputs from all model checkpoints post-RLAIF, our physician author noted that the ``Subjective'' sections were generally of high quality and nearly indistinguishable from notes authored by clinicians. However, the ``Assessment and Plan'' sections could be improved to more accurately and concisely reflect medical reasoning.

\subsection{Physician Reader Study}
Four internal medicine physicians and one family medicine physician, in a blinded review, evaluated notes authored by physicians, LLaMA-Clinic, and Gemini Pro based on three criteria: real-world readiness, completeness, and accuracy (Figure \ref{fig:main_result}A). The median word counts and interquartile ranges (IQR) for notes authored by physicians, LLaMA-Clinic, and Gemini Pro were 118 (IQR: 94-150), 128 (IQR: 108-145), and 128 (IQR: 100-164), respectively. No statistically significant differences in word counts were observed among the three groups (Kruskal-Wallis H test: \textit{p} = 0.292). We assessed inter-rater reliability (IRR) utilizing Gwet's AC2 statistics. The AC2 scores for the three metrics ranged from 0.80 to 0.82, signifying a high degree of agreement among reviewers.  

\begin{figure*}[ht]
    \captionsetup{labelfont=bf={bf},skip=12pt}
    \centering
    \includegraphics[width=1.3\textwidth]{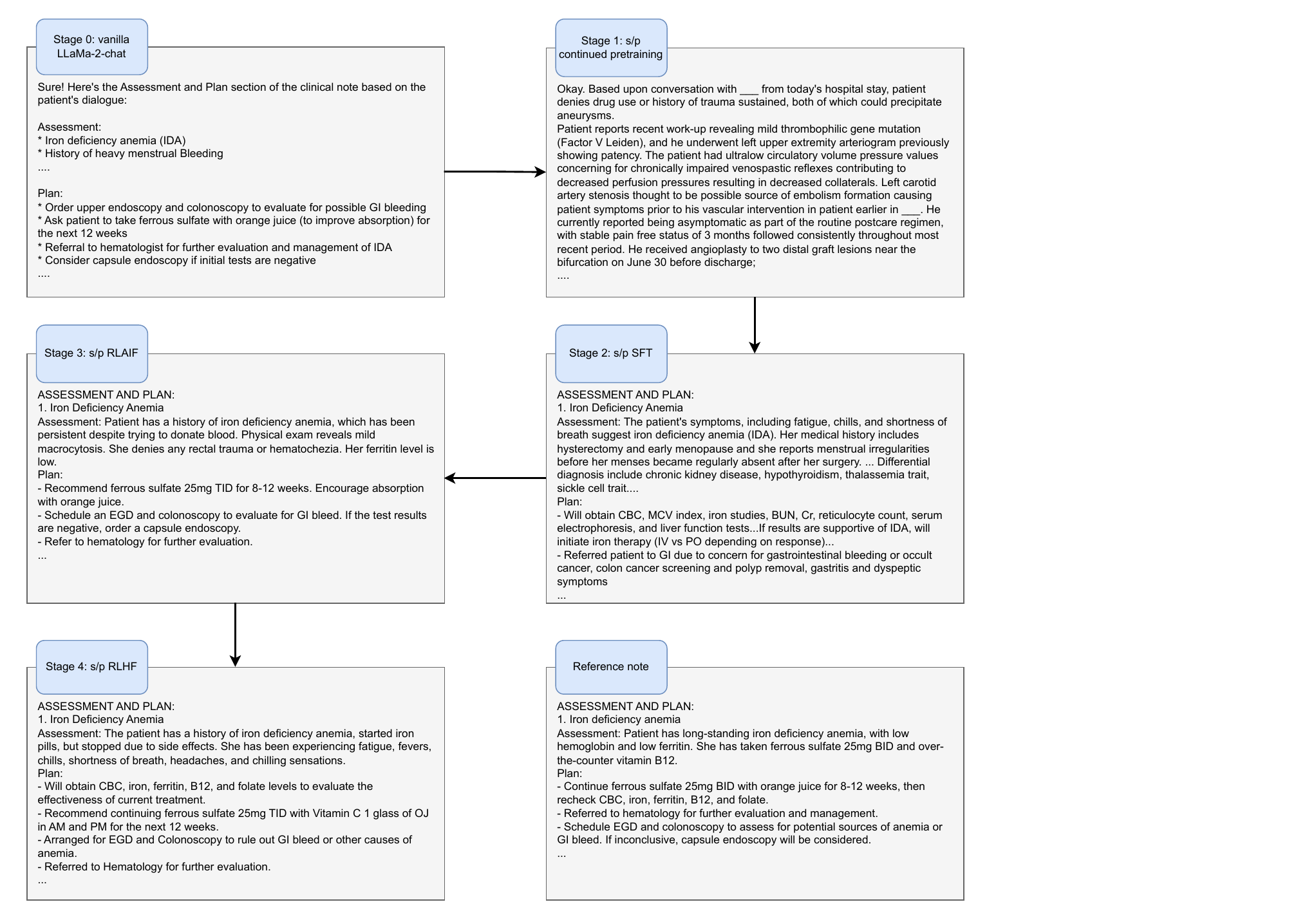}
    \renewcommand{\figurename}{Appendix Figure}
    \caption{\textbf{An Example of Model Outputs Progression with Training Steps.} Case number D2N073 from ACI-BENCH validation set. The figure illustrates the model's outputs for the same case following each training step. Continued pretraining adopted the style of discharge summaries but impaired the model's ability to follow instructions. Post-SFT, output quality improved, though it remained verbose. RLAIF and RLHF effectively refined outputs to match the format of reference notes and helped reduce hallucinations.}
    \label{fig:app.note_progression}
\end{figure*}

\begin{figure*}[!ht]
    \captionsetup{labelfont=bf={bf}}
    \centering
    \includegraphics[width=1\textwidth]{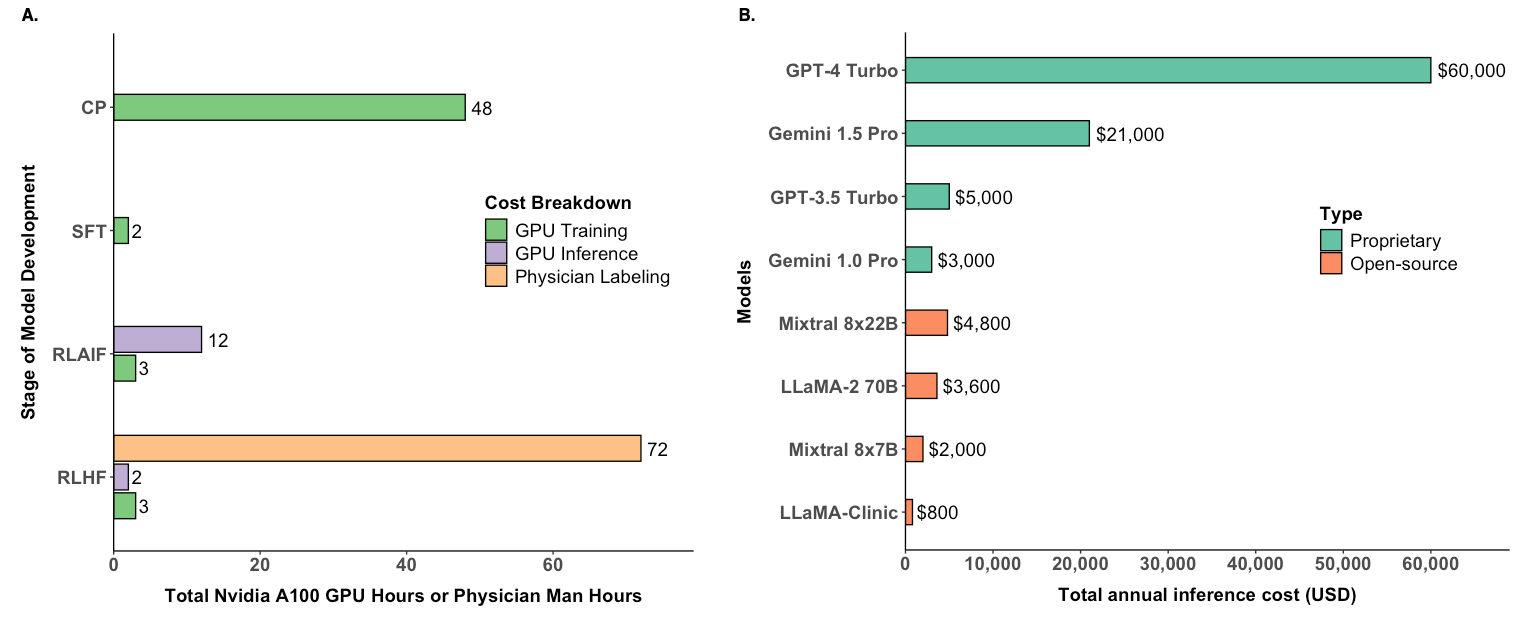}
    \renewcommand{\figurename}{Appendix Figure}
    \caption{\textbf{Cost Estimation for Model Development and Inference.} A. Model development cost for LLaMA-Clinic. The Y-axis of the horizontal bar chart represents different stages of model development, while the X-axis indicates costs, measured in total Nvidia A100 GPU hours or physician man hours. Importantly, the displayed costs only consider the training steps directly involved in developing LLaMA-Clinic and do not include the trial-and-error expenses from all experiments. B. Estimated inference costs for 1 million annual requests by proprietary and open-source models. The horizontal bar chart displays the annual inference cost estimation for both proprietary and open-source models, assuming 1 million requests for clinical note generation per year. The Y-axis represents the model names, and the X-axis indicates the total annual inference cost in US dollars. These calculations are based on the pricing per input and output tokens when using APIs on-demand. For proprietary models, pricing information was obtained from the websites of OpenAI and Google. For open-source models, pricing information was based on deploying models on Fireworks.ai with serverless inference. All pricing information was obtained in May 2024. In our study, Gemini 1.0 Pro served as the teacher model, while LLaMA-Clinic was the student model based on the LLaMA-2 13B.}
    
    \label{fig:app.cost}
\end{figure*}

\subsection{Cost Analysis for Model Development and Inference}
We provided our estimations of both GPU and human costs, measured in hours, for developing LLaMA-Clinic in Appendix Figure \ref{fig:app.cost}A. For inference cost, we calculated the cost of deploying open-source models in a serverless cloud environment provided by a third-party vendor. We compared these with the costs of using the proprietary models' APIs  (Appendix Figure \ref{fig:app.cost}B). Overall, proprietary models are more expensive than the open-source options. For proprietary models, costs significantly increase with the more advanced models. Similarly, for open-source models, costs increase with larger model sizes, as measured by the number of parameters. LLaMA-Clinic demonstrates a price advantage compared to its teacher model, Gemini 1.0 Pro, with a 3.75-fold cost reduction based on pricing information from May 2024. Assuming one million requests for clinical note generation, the estimated annual inference cost for LLaMA-Clinic is \$800 USD, compared to \$3,000 USD for Gemini 1.0 Pro.  Moreover, the amount mentioned is for the inference cost for one type of note. The total cost of supporting all types of notes will be significantly higher, but the relevant cost difference should remain the same.

\section{Additional Discussion}
\subsection{Continued Pretraining vs. Supervised Finetuning}
Continued pretraining of an LLM using a domain-specific corpus is recognized for enhancing performance on downstream tasks \cite{wu2023bloomberggpt}. This phase is considered a knowledge injection process, given that LLMs acquire the vast majority of their knowledge during the pretraining phase \cite{ovadia2023fine, zhou2023lima}. Several clinical LLMs that underwent continued pretraining with medical corpora, such as PubMed literature, combined with SFT have shown significant improvements in medical knowledge benchmarks \cite{chen2023meditron, wu2023pmcllama, luo2023biomedgpt}. 

However, a critical distinction exists between tasks focused on medical knowledge (e.g., answering USMLE questions) and those aimed at clinical note generation. Notably, GatorTronGPT \cite{peng2023study}, the only LLM trained from scratch using EMR data from real patients to date, performed lower in both MedQA and PubMedQA compared to other clinical LLMs \cite{chen2023meditron, wu2023pmcllama, kweon2023publicly, wang2024reflection}. This outcome indicates that EMR data alone may lack comprehensive medical knowledge. For the task of note generation, we hypothesized that continued pretraining with clinical notes could offer benefits by introducing greater lexical variance, unique semantic patterns, and diverse formatting similar to prior work \cite{lehman2023we}. 
Our experiments did not conclusively demonstrate the anticipated benefits of continued pretraining. Interestingly, the LLaMA-2 model without continued pretraining achieved the highest ROUGE-1 scores after SFT and RLAIF in our early experiments (see Appendix Table \ref{Table:RLHF2}). We opted to proceed with the continued pretrained model for RLHF due to subtle peculiarities observed upon manual inspection. Given the significant time and computational resources required for continued pretraining, its utility, particularly with clinical notes, merits further exploration in future work.

\subsection{Data Selection for Continued Pretraining}
Another potential factor in the less impressive improvement from continued pretraining may be attributed to the variance in data distribution between discharge summaries and outpatient notes. To address this, we performed experiments focused on a condensed version of discharge summaries (Discharge-short), hypothesizing that the ``brief hospital course'' section would contain data of higher quality than the complete discharge summary. Indeed, models trained on the Discharge-short outperformed those trained on the full summaries (Table \ref{table:s/p RLAIF}). Interestingly, we observe that during the pretraining stage, models trained on full summaries achieved lower training losses (see Figure \ref{fig:loss_curve}). However, this did not lead to better performance in the downstream task. We speculate that the structured nature of the full discharge summaries, which include sections such as laboratory results and medication lists, presents more straightforward learning targets for the model. This allows it to achieve lower training losses, which do not necessarily translate into improved task performance. This observation underscores the necessity for thorough analysis of the data used for pretraining.

\section{Prompts to Gemini Pro for Dialogue-G Creation}\label{app:propmpt_for_dialoguG}
We presented the prompt sent to Gemini Pro for generating dialogues of Dialogue-G in Appendix Table \ref{tab:dialogue_prompt}. 

\section{Prompts to Gemini Pro for Reference Note Generation}\label{app.prompt_fot_reference_note}
We presented the prompt used by Gemini Pro to create reference notes in Appendix Table \ref{tab:note_prompt}. We demonstrated two examples of clinical notes before and after the change in Appendix Figure \ref{fig:app.note_style}. 

\begin{figure*}[!ht]
    \captionsetup{labelfont=bf={bf},skip=12pt}
    \centering
    \includegraphics[width=\linewidth]{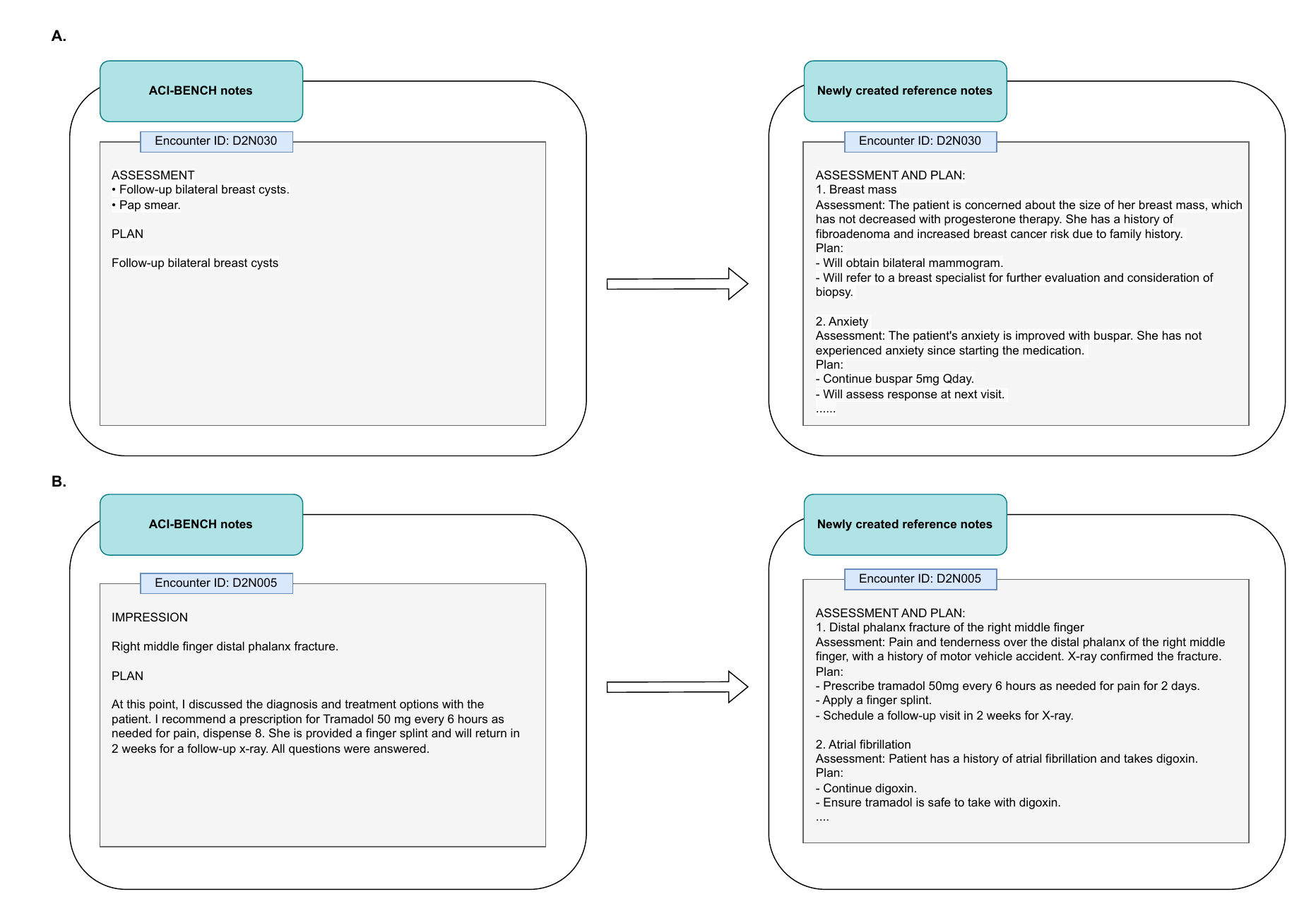}
    \renewcommand{\figurename}{Appendix Figure}
    \caption{\textbf{Comparison of Newly Created Reference Notes with ACI-BENCH Notes.} A. Example with the encounter ID D2N030. B. Example with the encounter ID D2N005. We compared newly generated reference notes using Gemini Pro with the original reference notes from two cases in ACI-BENCH. The newly created reference notes adhere more closely to our defined ``best practice'' format and contain more relevant medical information. }
    \label{fig:app.note_style}
\end{figure*}

\section{Instructions for Collecting Preference Data}\label{app:instruct_for_collect_preference}
We presented the instructions used for collecting physician preference data for RLHF in Appendix Table \ref{Table:preference}.

\section{Instructions and Scoring Rubric for Physician Reader Study}\label{app:instruct_for_physician_reader}
We presented the instructions used for the final physician reader study in Appendix Table \ref{Table:reader}.

\section{Instructions for Harm Evaluation}\label{app:instruct_for_harm}
We presented the instructions used for the harm evaluation in Appendix Table \ref{Table:harm}.

\section{An Example of Patient-Doctor Dialogue}\label{app.dialogue_example}
We presented an example of doctor-patient dialogue from ACI-BENCH in Appendix Table \ref{tab:dialogue}.  

\clearpage

\renewcommand{\arraystretch}{1.2}
\begin{table*}[ht]
    \small
    \captionsetup{labelfont=bf={bf}}
    \renewcommand{\tablename}{Appendix Table}

    \begin{tabular}{p{1.5cm} p{13cm}}
      
    \toprule
    \textbf{Category} & \textbf{Prompt}  \\ 
    \hline
    \textit{Dialogue} & Generate a synthetic patient-physician clinical dialogue encounter based on the clinical note below. Make sure all pertinent details are represented in the generated dialogue, so that a physician can easily write up the provided note. Pay close attention to make sure all details in the `ASSESSMENT AND PLAN', `HISTORY OF PRESENT ILLNESS' and `SUBJECTIVE' parts of the note are reflected in the dialogue. You may expand details to make the dialogue resemble a real clinical encounter. Denote doctor as [doctor] and patient as [patient]. \\
    \midrule
    \end{tabular}
    \caption{\textbf{Prompts to Gemini Pro for Dialogue-G Creation.}}
    \label{tab:dialogue_prompt}
\end{table*}

\renewcommand{\arraystretch}{1.2}
\begin{table*}[!ht]
    \small
    \captionsetup{labelfont=bf={bf}}
    \renewcommand{\tablename}{Appendix Table}
    \begin{tabular}{p{1.5cm} p{13cm}}
      
    \toprule
    \textbf{Category} & \textbf{Prompt}  \\ 
    \hline
    \textit{Subjective} & You are a physician writing a clinical note based on a dialogue with the patient. Only write the ``SUBJECTIVE'' part of note, which include the section of [CHIEF COMPLAINT] and [HISTORY OF PRESENT ILLNESS]. Only include information contained in the dialogue.  Follow the format as the example below:
\newline
\newline
    SUBJECTIVE 
    \newline
    \newline
    CHIEF COMPLAINT 
    \newline
    \newline
    Annual health maintenance examination. 
    \newline
    \newline
    HISTORY OF PRESENT ILLNESS 
    \newline
    \newline
    The patient is a pleasant [age]-year-old male who presents for his annual health maintenance examination. He reports no new complaints today. He denies any recent changes in his hearing. He continues to take niacin for his dyslipidemia, and he has had no problems with hemorrhoids in the last 6 months. He also denies any problems with concha bullosa of the left nostril or septal deviation. \\
    \midrule
    \textit{Assessment and Plan} & You are a physician writing a clinical note based on a dialogue with the patient. Only write the ``ASSESSMENT AND PLAN'' section of note. List each medical problem separately. Under each problem, include assessment (such as medical reasoning) and plan (both diagnostic and therapeutic ). At the end, may include a short section on follow up instruction when applicable. Only include information contained in the dialogue. Follow the format as the example below:
    \newline
    \newline
    ASSESSMENT AND PLAN: 
    \newline
    \newline
    1. Possible COPD exacerbation 
    \newline
    Assessment: Increased work of breathing with  wheezing on exam, suggesting COPD exacerbation. He does have frequent COPD exacerbation in the past. Differential diagnosis include pneumonia (though no fever or cough), PE (though no risk factors) or simple viral infection. 
     \newline
    Plan:
    \newline
    - WIll obtain CXR. 
    \newline
    - Will start duoneb therapy and oral prednisone 30mg Qday. 
    \newline
    \newline
    2. Hypertension 
    \newline
    Assessment: The patient's blood pressure is well controlled. 
    \newline
    Plan: 
    \newline
    - Continue lisinopril 20mg Qday. 
    \newline
    \newline
    Follow-up instructions: 
    \newline
    - return to clinic in 1 week, or sooner of failed to response with current treatment. \\
    \midrule
    \end{tabular}
  \caption{\textbf{Prompts to Gemini Pro for Reference Note Generation.}}
    \label{tab:note_prompt}
\end{table*}

\begin{table*}[ht]
\small
\renewcommand{\tablename}{Appendix Table}
\captionsetup{labelfont=bf={bf}}
\begin{tabular}{p{14.5cm}}
\toprule
\textbf{Instructions} \\
\midrule
\begin{enumerate}
    \item Please first read the dialogue and then pick your most and least preferred notes. Most conversation occurred in the outpatient setting. 
    \item We will only look at the ``Subjective'' and ``Assessment and Plan'' parts of a note. There will be a separate row for ``Subjective'' and ``Assessment and Plan'', respectively. 
    \item On each row, you will be given three notes generated by LLMs. Pick the MOST preferred note, and the LEAST preferred note by selecting the corresponding note number in the columns of “Preferred” and “Rejected”.
    \item Focus on whether the clinical note accurately reflected information from the conversation. Ignore any error related to medical knowledge, as long as the information was mentioned in the conversation.
    \item Base your preference on factors like clinical readiness, correctness, and adherence to the desired format, including:
   \begin{itemize}
      \item[a.] Clinical readiness: Is the note ready for clinical use and does it capture important information?
      \item[b.] Correctness: Does the note include less false information?
      \item[c.] Adherence to format: Does the ``Subjective'' section include ``Chief complaints'' and ``History of present illness''? Does the ``Assessment and Plan'' section list each problem separately and include ``Assessment:'' and ``Plan:'' with the required details?
   \end{itemize}
    \item Make your preference judgement from a clinician’s perspective, considering which note would be most/least helpful to you. 
\end{enumerate}
\\
\bottomrule
\end{tabular}
\caption{\textbf{Clinician Note Preference Instruction.}}
\label{Table:preference}
\end{table*}

\clearpage
\begin{table*}[ht]
\small
\captionsetup{labelfont=bf={bf}}
\renewcommand{\tablename}{Appendix Table}
\begin{tabular}{p{14.5cm}}
\toprule
\textbf{Instruction and rubrics} \\
\midrule
\begin{enumerate}
    \item In each row you will be given a synthetic outpatient patient-provider dialogue from ACI-BENCH, and three clinical notes based on the same dialogue. Two notes are generated by large language model, and one note is written by real physician. We have performed randomization of the notes (so that notes from the column of note\_1 are from different sources) and simple processing to unify the format of notes. 
    \item The dialogues from ACI-BENCH include conversations with (a) calls to a virtual assistant (b) unconstrained directions or discussions with a scribe, and (c) natural conversations between a doctor and patient.
    \item There are 40 dialogues. For each dialogue, we will evaluate ``Subjective'' and ``Assessment and Plan'' parts of the note in separate rows. Therefore, there are 80 rows in total. 
    \item For each row, you will first read the entire dialogue and then read the 3 notes. You will subsequently score the quality of each note for the 3 axes of ``accuracy'', ``completeness'' and ``readiness for real-world use''. For each axis, you will give a score of 1 to 5 (very poor, poor, acceptable, good, or very good), where higher number suggests better quality. It is OK to give the same score for different notes if you feel they are of similar quality. For each row since there are 3 notes, you will give total 9 scores. In each row there is a section of “Comment” for you to free text any feedback if you feel like to.
    \item \textbf{Accuracy}: For this axis, answer the question “does the factual information from clinical note accurately match that from the dialogue?” A note is accurate if it doesn’t say things that aren’t in the dialogue, it doesn’t mix up facts, and generally is not misleading. It might be acceptable if the note contains reasonable medical reasoning in the section of “Assessment”, for example in describing differential diagnosis. Please ignore any medical knowledge error, as long as the information was discussed in the dialogue.

\textbf{Rubric}:
\
\begin{itemize}
\item Score of 1 (very poor): The note contains a significant amount of content that is either factually incorrect, fabricated, or disconnected from the dialogue. 
\item Score of 3 (acceptable): The note contains some minor content that is either factually incorrect, fabricated, or disconnected from the dialogue.
\item Score of 5 (very good): The note has no incorrect statements or misleading implications.
\end{itemize}
    \item \textbf{Completeness}: For this axis, answer the question “how well does the note cover the important information from the dialogue?” An ideal clinical note would contain all clinically important information represented in the dialogue. Also, just as in real-world scenario, a good clinical note could be short but pertinent. A note has poor coverage if someone reading only the note would be missing several important pieces of information about the clinical encounter. Give your score based on what is typically expected from a clinical note. 
    
\textbf{Rubric}:
\
\begin{itemize}
\item Score of 1 (very poor): The note is missing a significant amount of important clinical information from the dialogue. 
\item Score of 3 (acceptable): The note is missing some minor piece of clinical information from the dialogue.
\item Score of 5 (very good): The note covers all important clinical information from the dialogue, as you would expect from a real-world note.
\end{itemize}
    \item \textbf{Readiness for real-world use}: For this axis, answer the question “which note is most ready for clinical use in the real-world scenario?” Answer this question imagine you are adopting AI-generated clinical notes for your everyday clinical work, and you will proofread and make edits to these notes before file into medical record. In this workflow, which note would you prefer the most? For example, this might be the note that meet your style and carries the most pertinent information without note bloating. Or a best note for you might be the one that requires the least amounts of edits from you (even if it contains some minor error). In other words, you can think of this as scoring the overall quality of the note for the workflow.   
    
\textbf{Rubric}:
\begin{itemize}
\item Score of 1(very poor): The note is impossible to use or would require significant edits from you.
\item Score of 3 (acceptable): The note requires some edits from you.
\item Score of 5 (very good): The note is ready for clinical use without any further edits from you.
\end{itemize}
\end{enumerate}
\\
\bottomrule
\end{tabular}
\caption{\textbf{Instructions for Physician Reader Study.}}
\label{Table:reader}
\end{table*}

\clearpage

\begin{table*}[ht]
\small
\captionsetup{font={bf}}
\renewcommand{\tablename}{Appendix Table}
\begin{tabular}{p{14.5cm}}
\toprule
\textbf{Instructions} \\
\midrule
For those notes that you gave a score of less than 5 for ``Accuracy'' or ``Completeness'', please answer two questions:
Suppose the note is used in the standard clinical workflow, what would be: 

\begin{enumerate}
    \item  \textbf{“... extent of possible harm?”} 
    
    \textbf{Rubric}:
\begin{itemize}
\item Score of 1: None.
\item Score of 2: Mild or moderate harm.
\item Score of 3: Severe harm or death.
\end{itemize}

    \item \textbf{“... likelihood of possible harm?”}
    
    \textbf{Rubric}:
\begin{itemize}
\item Score of 1: Low.
\item Score of 2: Medium.
\item Score of 3: High.
\end{itemize}

\end{enumerate}
\\
\bottomrule
\end{tabular}
\caption{\textbf{Harm Evaluation Instruction.}}
\label{Table:harm}
\end{table*}

\begin{table*}[h]
\small
\captionsetup{labelfont=bf={bf}}
\renewcommand{\tablename}{Appendix Table}
\renewcommand{\arraystretch}{1.5}
\begin{tabular}{llcc}
\cline{1-4}
\textbf{Type} & \textbf{Models} & \textbf{Cost / 1 Million Input Tokens (USD)} & \textbf{Cost / 1 Million Output Tokens (USD)} \\
\cline{1-4}
\multirow{4}{*}{Proprietary} & Gemini 1.5 Pro & 3.5 & 10.5 \\
                             & Gemini 1.0 Pro & 0.5 & 1.5 \\
                             & GPT-4 Turbo    & 10.0 & 30.0 \\
                             & GPT-3.5 Turbo  & 1.0 & 2.0 \\
\cline{1-4}
\multirow{3}{*}{Open-Source} & LLaMA-Clinic & 0.2 & 0.2 \\
                             & LLaMA-2 70B  & 0.9 & 0.9 \\
                             & Mixtral 8x7B & 0.5 & 0.5 \\
                             & Mixtral 8x22B & 1.2 & 1.2 \\
\cline{1-4}
\end{tabular}
\caption{\textbf{Pricing Information for Inference Cost Estimation.} We sourced pricing information from the websites of Google AI \cite{GoogleAIPricing2024}, OpenAI \cite{OpenAIAPIPricing2024}, and Fireworks.ai \cite{FireworksAI2024} in May 2024 for on-demand API calls. The price of GPT-3.5 Turbo is based on the model gpt-3.5-turbo-1106. Fireworks.ai charges the same price for both input and output tokens.}
\label{tab:token_costs}
\end{table*}

\clearpage

\begin{table*}[ht]
\small
\renewcommand{\tablename}{Appendix Table}
\captionsetup{labelfont=bf={bf}}
\begin{tabular}{p{14.5cm}}
\toprule
\textbf{Dialogue} \\
\midrule
\noindent \textbf{Doctor:} Hi, Martha. How are you? \\
\textbf{Patient:} I'm doing okay. How are you? \\
\textbf{Doctor:} I'm doing okay. So, I know the nurse told you about DAX. I'd like to tell DAX a little bit about you, okay? \\
\textbf{Patient:} Okay. \\
\textbf{Doctor:} Martha is a 50-year-old female with a past medical history significant for congestive heart failure, depression, and hypertension who presents for her annual exam. So, Martha, it's been a year since I've seen you. How are you doing? \\
\textbf{Patient:} I'm doing well. I've been traveling a lot recently since things have, have gotten a bit lighter. And I got my vaccine, so I feel safer about traveling. I've been doing a lot of hiking. Uh, went to Washington last weekend to hike in northern cascades, like around the Mount Baker area. \\
\textbf{Doctor:} Nice. That's great. I'm glad to hear that you're staying active, you know. I just love this weather. I'm so happy the summer is over. I'm definitely more of a fall person. \\
\textbf{Patient:} Yes, fall foliage is the best. \\
\textbf{Doctor:} Yeah. Um, so tell me, how are you doing with the congestive heart failure? How are you doing watching your diet? I know we've talked about watching a low sodium diet. Are you doing okay with that? \\
\textbf{Patient:} I've been doing well with that. I resisted, as much, as I could, from the tater tots, you know, the soft pretzels, the salty foods that I, I love to eat. And I've been doing a really good job. \\
...\\
\textbf{Doctor:} Hey, Dragon, show me the blood pressure. So, yeah, looking at your blood pressure today here in the office, it is a little elevated. You know, it could just, you could just be nervous. Uh, let's look at some of the past readings. Hey, Dragon, show me the blood pressure readings. Here we go. Uh, so they are running on the higher side. Um, y- you know, I, I do think that, you know, I'd like to see you take your medication a little bit more, so that we can get that under control a little bit better, okay? \\
\textbf{Patient:} Okay. \\
\textbf{Doctor:} So, I'm just gonna check out your heart and your lungs. And you know, let you know what I find, okay? \\
\textbf{Patient:} Okay. \\
\textbf{Doctor:} Okay. So, on your physical examination, you know, everything looks good. On your heart exam, I do appreciate a three out of six systolic ejection murmur, which I've heard in the past, okay? And on your lower extremities, I do appreciate one plus pitting edema, so you do have a little bit of fluid in your legs, okay? \\
\textbf{Patient:} Okay. \\
\textbf{Doctor:} Let's go ahead, I wanna look at some of your results, okay? Hey, Dragon, show me the echocardiogram. So, this is the result of the echocardiogram that we did last year. It showed that you have that low-ish pumping function of your heart at about 45\%. And it also shows some mitral regurgitation, that's that heart murmur that I heard, okay? \\
...\\
\textbf{Doctor:} Um, so I wanna just go over a little bit about my assessment and my plan for you, okay? So, for your first problem your congestive heart failure, um, I wanna continue you on your current medications. But I do wanna increase your lisinopril to 40 milligrams a day, just because your blood pressure's high. And you know, you are retaining a little bit of fluid. I also wanna start you on some Lasix, you know, 20 milligrams a day. And have you continue to watch your, your diet, okay? \\
\textbf{Patient:} Okay. \\
\textbf{Doctor:} I also wanna repeat another echocardiogram, okay? \\
\textbf{Patient:} All right. \\
\textbf{Doctor:} Hey, Dragon, order an echocardiogram. From a depression standpoint, it sounds like you're doing really well with that. So, I'm, I'm really happy for you. I'm, I'm glad to see that you're in therapy and you're doing really well. I don't feel the need to start you on any medications this year, unless you feel differently. \\
\textbf{Patient:} No, I feel the same way. \\
...\\

\bottomrule
\end{tabular}
\caption{\textbf{An Example of Patient-Doctor Dialogue from ACI-BENCH with Encounter ID D2N001}. We corrected minor grammatical and spelling errors for display purposes.}
\label{tab:dialogue}
\end{table*}

\end{document}